\documentclass[10pt,twocolumn,letterpaper]{article}

\usepackage[pagenumbers]{cvpr} 

\usepackage{booktabs}
\usepackage{makecell}
\usepackage{xspace}
\usepackage{adjustbox}   

\newcommand{\name}{\textbf{SOC}\xspace}

\newcommand{\od}{\textbf{OD}\xspace}
\newcommand{\seg}{\textbf{SEG}\xspace}

\newcommand{\vg}{\textbf{VG}\xspace}

\newcommand{\fc}{\textbf{\name-FC}\xspace}
\newcommand{\gc}{\textbf{\name-GC}\xspace}

\newcommand{\sfc}{\textbf{\name-SFC}\xspace}
\newcommand{\sgc}{\textbf{\name-SGC}\xspace}

\newcommand{\mixcoco}{\textbf{\name-COCO-Mix}\xspace}
\newcommand{\namelvis}{\textbf{\name-LVIS-Category}\xspace}

\newcommand{\fcfiftyk}{\textbf{FC-50K}\xspace}

\definecolor{cvprblue}{rgb}{0.21,0.49,0.74}
\usepackage[pagebackref,breaklinks,colorlinks,allcolors=cvprblue]{hyperref}

\def\paperID{*****} 
\def\confName{CVPR}
\def\confYear{2026}

\usepackage[utf8]{inputenc} 
\usepackage[T1]{fontenc}    
\usepackage{hyperref}       
\usepackage{url}            
\usepackage{booktabs}       
\usepackage{amsfonts}       
\usepackage{nicefrac}       
\usepackage{microtype}      
\usepackage{xcolor}         
\usepackage{algorithm}
\usepackage{algorithmicx}
\usepackage[noend]{algpseudocode}
\usepackage{enumitem}   %
\usepackage{graphicx}
\usepackage{amsmath}
\usepackage{multirow}
\usepackage{booktabs}
\usepackage{tabularx}
\usepackage{adjustbox}
\usepackage{caption}
\usepackage{subcaption}
\makeatletter
\@namedef{ver@eso-pic.sty}{2020/10/14}
\makeatother
\usepackage{pdfpages}
\usepackage{pifont}
\usepackage{afterpage}
\newcommand{\cmark}{\ding{51}}
\newcommand{\xmark}{\ding{55}}

\title{Synthetic Object Compositions for Scalable and Accurate \\ Learning in Detection, Segmentation, and Grounding}
\author{%
  Weikai Huang$^{1}$
  Jieyu Zhang$^{1}$
  Taoyang Jia$^{1}$
  Chenhao Zheng$^{1}$\\
  Ziqi Gao$^{1}$
  Jae Sung Park$^{1}$
  Winson Han$^{2}$
  Ranjay Krishna$^{1,2}$
  \\[0.3em] 
  $^{1}$University of Washington \quad
  $^{2}$Allen Institute for Artificial Intelligence
  \\[1em]
}
\begin{document}
\twocolumn[{
\maketitle
\begin{center}
\includegraphics[width=\textwidth]{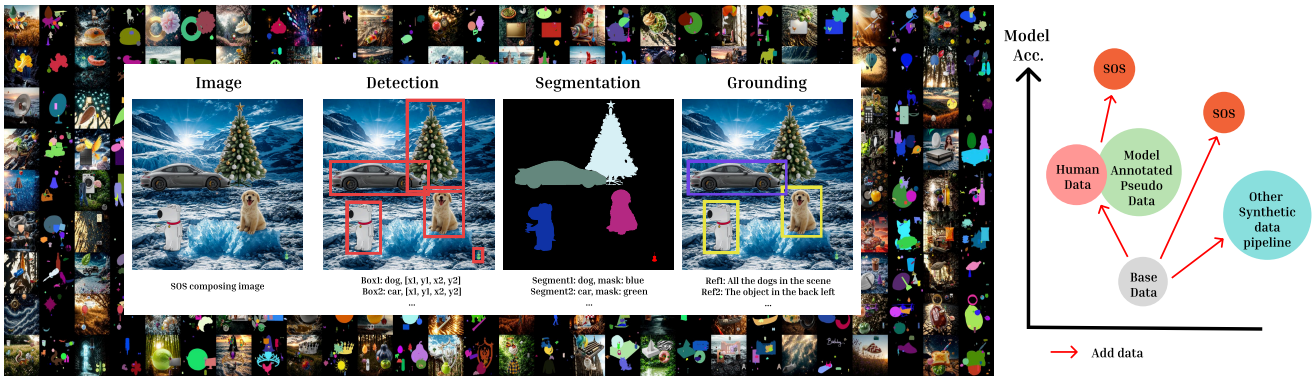}
\vspace{-18pt}
\captionof{figure}{\textbf{\name data examples.} With \name, we generate 20M object segments and compose them into 2M images via 3D geometric layout augmentation, producing accurate and diverse masks, bounding boxes, and referring expressions for object detection, instance segmentation, and visual grounding. Right: \name surpasses both real and synthetic data generation pipelines and complements real datasets when combined.}
\label{fig:teaser}
\end{center}
}]

\begin{abstract}
Visual grouping—operationalized through tasks such as instance segmentation, visual grounding, and object detection—enables applications ranging from robotic perception to photo editing. These fundamental problems in computer vision are powered by large-scale, painstakingly annotated datasets.
Despite their impact, these datasets are costly to build, biased in coverage, and difficult to scale. 
Synthetic datasets offer a promising alternative but struggle with flexibility, accuracy, and compositional diversity.
We introduce \name, an accurate and scalable data synthesis pipeline via a novel object-centric composition strategy.
It composes high-quality synthetic object segments into new images using 3D geometric layout augmentation and camera configuration augmentation with generative harmonization and mask-area-weighted blending, yielding accurate and diverse masks, boxes, and referring expressions.
Models trained on just 100K of our synthetic images outperform those trained on larger real datasets (GRIT 20M, V3Det 200K) and synthetic pipelines (Copy-Paste, X-Paste, SynGround, SegGen by +24-36\%)—achieving +10.9 AP on LVIS and +8.4 N\textsubscript{Acc} on gRefCOCO. \name\ also enables controllable dataset construction for different use cases and boosts performance in both low-data and closed-vocabulary scenarios. Augmenting LVIS and COCO with synthetic object segments delivers strong performance across different real data scales and yields even greater improvements when real data is extremely limited (+6.59 AP on 1\% COCO data). Furthermore, this controllability enables targeted data generation for intra-class referring, a diagnostic grounding task we propose that requires fine-grained attribute discrimination.
\end{abstract}

\section{Introduction}

Human perception is inherently structured: we naturally organize the visual world into coherent regions and objects based on principles of proximity, similarity, continuity, and closure—phenomena first formalized in Gestalt psychology~\cite{wertheimer1923untersuchungen, kanizsa1979organization}. 
In computer vision, tasks such as object detection, instance segmentation, and referring expression grounding operationalize this capability and power a broad range of applications, from robotic perception and autonomous driving to photo editing and assistive technologies.

Historically, the success of segmentation and detection models remains heavily dependent on the quality, scale, and diversity of annotated datasets~\cite{Lin2014COCO,Cordts2016Cityscapes,Gupta_2019_CVPR}.
Existing annotated datasets are painstaking and expensive to expand.
For example, COCO~\cite{lin2015microsoftcococommonobjects} required 2.2M worker hours to annotate only 100K  images with 80 object categories.
They are also biased toward a limited number of categories, and often fail to cover the full combinatorial diversity of scenes, objects, and referring expressions encountered in the real world.


To overcome these challenges, researchers have turned to synthetic data. Synthetic datasets by rendering scenes in simulation~\cite{Jiang2022PseudoQGP,Peng2023Kosmos2GM,Wang2024LearningVG,he2024learningsyntheticdatavisual,zhang2025scaling,kirillov2023segment,wu2023datasetdmsynthesizingdataperception,yang2023freemasksyntheticimagesdense,ye2024seggensuperchargingsegmentationmodels,qian2024maskfactoryhighqualitysynthetic,Wrenninge2018Synscapes,Cabon2020VKITTI2,Roberts_2021_ICCV,Liu2019CLEVRRef} allow for full control and offer accurate dense ground truth at a large scale, but are still geared toward rigid domains like indoor or driving scenes and are limited in object diversity due to the scarcity of 3D object assets. On the other hand, model-annotated synthetic datasets on real or generated images~\cite{Jiang2022PseudoQGP,Peng2023Kosmos2GM,Wang2024LearningVG,he2024learningsyntheticdatavisual,zhang2025scaling,kirillov2023segment,wu2023datasetdmsynthesizingdataperception,yang2023freemasksyntheticimagesdense,ye2024seggensuperchargingsegmentationmodels,qian2024maskfactoryhighqualitysynthetic} introduce greater variability in scene composition and object appearance, yet they inherit annotation noise from both the labeling models and the image generators.

In this work, we introduce \name (Figure~\ref{fig:teaser}): a scalable data synthesis pipeline that departs from prior paradigms.
Rather than rendering entire scenes or relying on models to annotate images, we employ an object-centric composition strategy: we first collect high-quality synthetic object segments using a strong generative model; next, we compose new images by pasting segments according to designed layouts with 3D geometric layout augmentation and camera configuration augmentation to ensure diversity and robustness; finally, we apply generative harmonization with mask-area-weighted blending leveraging recent advances~\cite{zhang2025scaling} to enhance the images' photorealism.
\name not only provides controllability and accurate region annotations like simulation-based methods, but also offers the composition flexibility and scene diversity of model-annotated approaches.
Specifically, we generate 20M synthetic object segments; using these, one can compose any number of synthetic images, each with corresponding mask, box, label, and referring expression annotations.





In open‐vocabulary detection and visual grounding, we show that \name\ is an efficient and scalable approach to improve model performance, and complementary to real datasets. Even a small \name\ dataset of 50K images delivers strong gains: it lifts LVIS~\cite{gupta2019lvisdatasetlargevocabulary} AP from 20.1 to 29.8 (+9.7) and rare‐class AP from 10.1 to 23.5 (+13.4), exceeding the +7.0 boost from 20M model generated data (GRIT~\cite{Wu2022GRiTAG}) and matching the +10.5 gain from 200K human-annotated data (V3Det~\cite{zhao2024open,Wu2022GRiTAG,wang2023v3detvastvocabularyvisual}). Scaling \name from 50K to 400K further boosts LVIS AP to 31.4 (+1.6) and OdinW‐35 mAP to 22.8 (+1.8), demonstrating the scalability of \name. Additionally, \name complements existing real datasets with more balanced category distribution, wider coverages, and compositions: adding 100K \name images above model-generated data (GRIT) and human-annotated data (V3Det) still yields additive gains of +6.2 rare‐class LVIS AP, and +2.8 average AP on OdinW‐35. 

Next, augmenting instance segmentation datasets with \name\ delivers consistent improvements across different data regimes. Incorporating 50K \name\ images increases LVIS AP$_\mathrm{rare}$ from 40.87 to 44.70 (+3.83) and overall AP from 46.96 to 48.48 (+1.52) compared to training on LVIS alone. In COCO, mixing synthetic \name\ segments brings an average 3\% performance boost, and yields a +6.59 AP gain in an extremely low-data setup when the model is trained on only 1\% of COCO images. These results demonstrate that synthetic segments not only complement but amplify the real annotations in extremely limited‐data regimes.

Compared to other synthetic data generation pipelines, \name substantially outperforms Copy-Paste (+36.1\% on COCO instance segmentation), X-Paste (+36.0\% on COCO), SynGround (+24.1\% on COCO), and SegGen (+28.5\% on COCO), demonstrating the effectiveness of our object-centric composition approach with 3D geometric layout augmentation and generative harmonization.

Finally, the controllability of \name\ enables targeted improvements on the intra‐class referring task, a diagnostic variant of the referring expression task where the model needs to distinguish objects of the same category but with different attributes. By generating a \name dataset specifically with multiple objects of the same category but different attributes, we bring greater model performance increase than existing datasets (namely, GRIT or V3Det).

To summarize, our contributions are three-fold: (1) We release \name, a large-scale synthetic dataset comprising 20M object segments and 2M composed images with pixel-perfect annotations, representing the first large synthetic dataset to surpass real datasets across diverse models and tasks. (2) We demonstrate that \name outperforms all existing synthetic data pipelines and complements real datasets across multiple benchmarks spanning object detection, instance segmentation, and visual grounding tasks. (3) We propose a new intra-class referring expression benchmark with human annotations and show that \name's controllability enables targeted data synthesis to effectively address this diagnostic task.

\section{Related work}
We position our contribution within the landscape of datasets and synthetic data generation methods for object detection, segmentation, and visual grounding. Detailed related work is provided in the Appendix.

\noindent \textbf{Datasets for object detection, segmentation, and referring expression.}
Detection and segmentation benchmarks include COCO~\cite{lin2015microsoftcococommonobjects}, OpenImages~\cite{Kuznetsova_2020}, Object365~\cite{9009553}, LVIS~\cite{gupta2019lvisdatasetlargevocabulary}, V3Det~\cite{wang2023v3detvastvocabularyvisual}, and ODInW~\cite{li2021grounded, li2022elevater}. Semantic segmentation datasets include COCO‐Stuff~\cite{Caesar2016COCOStuffTA}, ADE20K~\cite{8100027, zhou2018semanticunderstandingscenesade20k}, PASCAL VOC~\cite{everingham2011pascal}, PASCAL Context~\cite{Mottaghi2014TheRO}, and Cityscapes~\cite{Cordts2016Cityscapes}. Referring expression benchmarks include ReferItGame~\cite{kazemzadeh-etal-2014-referitgame}, RefCOCO/+/g~\cite{Yu2016ModelingCI, Mao2015GenerationAC}, Flickr30K Entities~\cite{plummer2016flickr30kentitiescollectingregiontophrase}, Visual Genome~\cite{Krishna2016VisualGC}, and GoldG~\cite{li2022groundedlanguageimagepretraining}.

\noindent \textbf{Synthetic data generation pipelines.}
Copy-paste methods~\cite{Dvornik2018ModelingVC, Fang2019InstaBoostBI, Ghiasi2020SimpleCI} and X-Paste~\cite{Zhao2022XPasteRS} paste segments onto backgrounds but lack photorealism. Diffusion-based approaches including ControlNet~\cite{Zhang2023AddingCC}, LayoutDiffusion~\cite{Zheng2023LayoutDiffusionCD}, MaskFactory~\cite{qian2024maskfactoryhighqualitysynthetic}, SegGen~\cite{ye2024seggensuperchargingsegmentationmodels}, and DatasetDM~\cite{wu2023datasetdmsynthesizingdataperception} generate images from masks or layouts but often produce inaccurate annotations. Pseudo-labeling methods including GRIT~\cite{gupta2022gritgeneralrobustimage}, SynGround~\cite{he2024learningsyntheticdatavisual}, and Learning VG~\cite{Wang2024LearningVG} apply detectors to real or generated images. In contrast, \name uses object composition with 3D geometric layout augmentation and generative harmonization for pixel-perfect annotations.

\noindent \textbf{Image matting, blending, and harmonization.}
Image matting methods including Deep Image Matting~\cite{Xu2017DeepIM}, attention-guided matting~\cite{Qiao2020AttentionGuidedHS}, Human Instance Matting~\cite{Sun2022HumanIM}, Referring Image Matting~\cite{Li2023ReferringIM}, DFIMat~\cite{Jiao2024DFIMatDF}, MP-Mat~\cite{Jiao2025MPMatA3H}, and DIS~\cite{qin2022highlyaccuratedichotomousimage} extract object segments with precise alpha mattes. We use DIS for segment extraction and IC-Light~\cite{zhang2025scaling} for diffusion-based harmonization, combined with mask-area-weighted blending for photorealism.

\begin{figure*}[!t]
    \centering
    \includegraphics[width=0.8\linewidth]{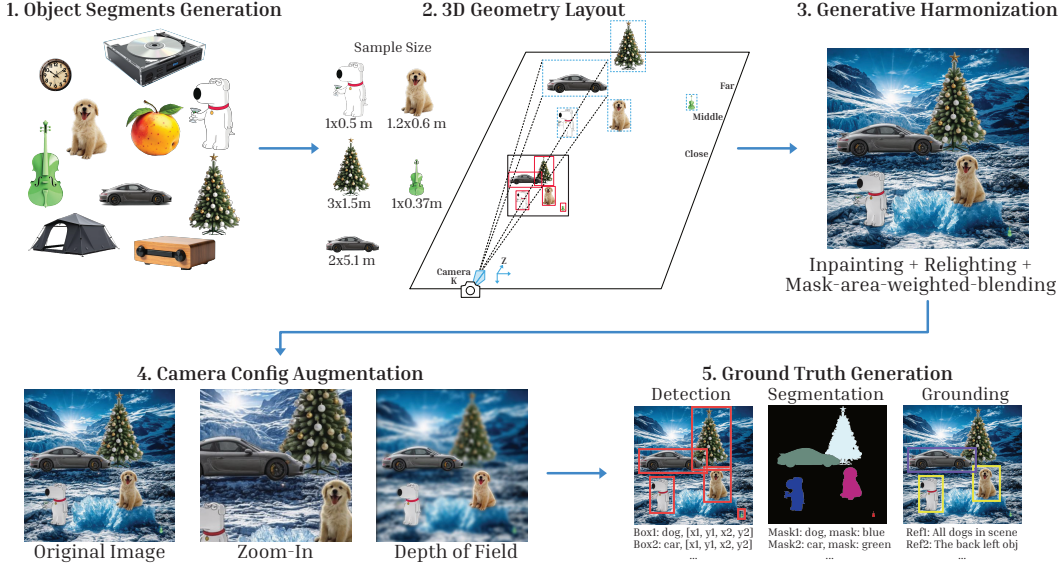}
    \vspace{-0.5em}
    \caption{\textbf{Overview of \name\ Pipeline.}
      (1) \textbf{Object Segments Generation.} Generating object segments in diverse categories.
      (2) \textbf{3D Geometric Layout Augmentation.} Sample and place 5–20 segments per image with category-independent 3D scene modeling.
      (3) \textbf{Generative Harmonization.} Apply inpainting, global relighting, and mask-area-weighted blending to enhance realism and prevent models from learning only the edge without semantics.
      (4) \textbf{Camera Configuration Augmentation.} Apply random scaling, cropping, and depth-of-field blur to simulate diverse camera configurations.
      (5) \textbf{Generating Region Annotations.} Compute final masks, bounding boxes, and dense referring expressions.
    }
    \label{fig:pipeline}
    \vspace{-1em}
  \end{figure*}
\section{Method}

\name introduces a simple yet scalable pipeline for creating object detection (OD), instance segmentation (SEG), and visual grounding (VG) datasets synthetically.
Whereas existing methods begin with real or synthetic images and then rely on human annotators or automated models to generate bounding boxes, masks, or grounding labels, our pipeline (Figure~\ref{fig:pipeline}) assembles scenes from the ground up. We first build a massive library of object segments, then composite them into images—automatically producing region annotations without post-hoc image labeling. We then apply generative harmonization and image blending algorithms to provide images without obvious artifacts.

\subsection{Synthesizing object segments}
Given 46000+ object categories we collected, we and prompt Qwen 2.5-32B~\citep{qwen2025qwen25technicalreport} to produce text descriptions for each of them.
These prompts are then passed to FLUX‑1‑dev~\citep{flux2024}, a strong text-to-image generation model, yielding single‑object images rendered on a uniform white background with randomly sampled viewpoints to achieve diverse visual info for each objects..
Finally, we apply DIS~\citep{qin2022highlyaccuratedichotomousimage} to extract object segments from white-background images.
We find that this setup produces cleaner mask boundaries than generating objects in cluttered scenes.

\subsection{3D geometric layout augmentation}

Models trained on real datasets often exploit spurious correlations between object categories and pictorial cues (e.g., ``cars appear large and near the bottom'')~\cite{chen2020monopair}, rather than learning robust semantic features. To break these shortcuts, we adopt a \textit{3D geometric layout augmentation} strategy: we model each composite image as a 3D scene where \textit{depth and spatial position are sampled independently of object category}, i.e., $p(d_i, X_i, Y_i | c_i) = p(d_i, X_i, Y_i)$. This ensures objects of the same category appear at diverse depths, sizes, and positions, preventing category-specific pictorial patterns. For each image, we sample 5-20 object segments (matching COCO/SA-1B distributions) using balanced category sampling to avoid bias.

\noindent\textbf{3D scene modeling with category-independent sampling.}
Each object category $c$ has a commonsense physical size range (e.g., cars: 4-5m, cups: 10-20cm) generated by Qwen2.5-32B. We first sample camera focal length $f \sim \mathcal{U}(f_{\min}, f_{\max})$ and define a maximum depth $D_{\max} = \alpha \cdot f$ (where $\alpha$ is a scaling constant) that scales with the focal length to ensure appropriate scene coverage. We then define three depth ranges (in meters): close $[0.1 D_{\max}, 0.3 D_{\max}]$, middle $[0.3 D_{\max}, 0.6 D_{\max}]$, and far $[0.6 D_{\max}, D_{\max}]$. For each object segment $i$ of category $c_i$, we sample its physical size $S_i \sim \mathcal{N}(\mu_{c_i}, \sigma_{c_i})$ and depth $d_i$ (in meters) from one of the three ranges, following the depth distribution observed in COCO/SA-1B (40\% close, 35\% middle, 25\% far). We sample 3D position $(X_i, Y_i) \sim \mathcal{U}(X_{\min}, X_{\max}) \times \mathcal{U}(Y_{\min}, Y_{\max})$ uniformly (in meters) and project to 2D via perspective projection:
\begin{equation}
x_i = f \cdot \frac{X_i}{d_i}, \quad y_i = f \cdot \frac{Y_i}{d_i}, \quad s_i = f \cdot \frac{S_i}{d_i}
\end{equation}
where $(x_i, y_i)$ is the 2D center position and $s_i$ is the apparent size in pixels. We enforce constraints on the projected objects: if an object's apparent size is too small or too large, or if it completely occludes another object ($\text{IoU}(M_i, M_j) \geq 0.9$), we resample its 3D position and depth until valid placement is achieved. Our ablation (Section~\ref{exp:mask2former_ablation_10k}) validates this design: 10.03 AP vs. 8.60 AP (COCO layout) and 9.07 AP (random 2D layout).

\subsection{Generative harmonization}
\begin{figure}[!h]
  \centering
  \includegraphics[width=\columnwidth]{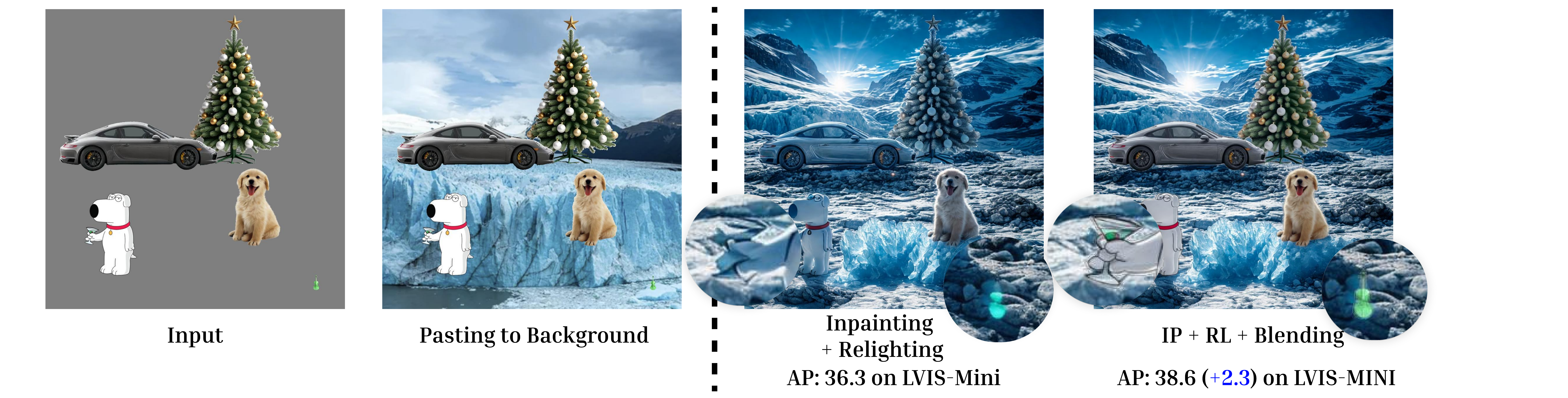}
    \caption{\textbf{ Comparison of Generative Harmonization Strategies.} From left to right: (1) input foreground images, (2) naively pasted onto a background, (3) Inpainting and Inpainting and global relighting with IC-Light (AP = 36.3 on LVIS-Mini), and (4) Inpainting plus relintiingng plus relighting plus mask-area-weighted blending (AP = 38.6 (+2.3) on LVIS-Mini). Blending notably preserves fine details of small objects and color fidelity and improves model performance.}
  \label{fig:relight-comparison}
  \vspace{-1em}
\end{figure}

Empirically, we find that naively pasting objects onto a background image often produces unnaturally sharp edges; segmentation models trained on these data can exploit these edge artifacts rather than truly learning semantics. To mitigate this shortcut, we employ IC‑Light~\citep{zhang2025scaling}, a diffusion‑based model that simultaneously performs background inpainting and global relighting to harmonize the composition. IC-Light generates a compatible background for the pasted objects and adjusts lighting conditions across the entire scene, producing more photorealistic images without simple edge‑based artifacts. However, IC-Light sometimes distorts fine details of smaller objects or alters object colors (e.g. blue $\rightarrow$ red), breaking consistency with text descriptions. To address this, we re-blend the original segments with the harmonized images using a blending weight $\alpha_i \in [0,1]$ for each object mask $M_i$, where smaller objects receive higher $\alpha_i$ to preserve their original appearance. Finally, we apply a lightweight soft matting step that converts the binary masks into soft alpha mattes to smooth object boundaries and better integrate the foregrounds into the harmonized background. This blending procedure yields a 2.3\% AP boost on LVIS-mini-val (Fig.~\ref{fig:relight-comparison}).

\begin{table*}[!t]
  \caption{\textbf{Qualitative comparison of other data generation approaches.} We categorize existing methods by their data generation paradigm and compare key capabilities. \name uniquely combines accurate pixel-level annotations, fine-grained controllability, multi-object composition, and open-vocabulary coverage at scale. Quantitative comparison with representative methods from each category is shown in each subsection. }
  \vspace{-1em}
  \centering
  \resizebox{0.95\textwidth}{!}{
    \begin{tabular}{llcccccl}
      \toprule
      \textbf{Method Category} &
      \textbf{Example Methods} &
      \textbf{Multi-object} &
      \textbf{Accurate Annotation} &
      \textbf{Controllability} &
      \textbf{Open-vocab} &
      \textbf{Scale} &
      \textbf{Compared in} \\
      \midrule
      Human-annotated
        & COCO~\cite{lin2015microsoftcococommonobjects}, LVIS~\cite{gupta2019lvisdatasetlargevocabulary}, V3Det~\cite{wang2023v3detvastvocabularyvisual}
        & \cmark
        & \cmark
        & \xmark
        & Limited
        & $\sim$1M
        & Sec~\ref{exp:ovd}--\ref{exp:intra-class} \\
      \midrule
      Real Image + Pseudo annotator
        & GRIT~\cite{gupta2022gritgeneralrobustimage}
        & \cmark
        & \xmark~(bbox only)
        & \xmark
        & \cmark
        & $\sim$20M
        & Sec~\ref{exp:ovd}, \ref{exp:vg}, \ref{exp:intra-class} \\
      \midrule
      Diffusion Image + Pseudo annotator
        & SynGround~\cite{he2024learningsyntheticdatavisual}
        & \cmark
        & \xmark~(pseudo)
        & Limited
        & \cmark
        & Unlimited (current 1.2M)
        & Sec~\ref{exp:compare_synthetic} \\
      \midrule
      Mask/Layout control Diffusion Image
        & \makecell[l]{MaskFactory~\cite{qian2024maskfactoryhighqualitysynthetic}, SegGen~\cite{ye2024seggensuperchargingsegmentationmodels}}
        & Limited
        & Limited
        & Limited
        & \cmark
        & Unlimited (current 1M)
        & Sec~\ref{exp:compare_synthetic} \\
      \midrule
      Simple-Copy-Paste
        & Simple-Copy-Paste~\cite{Ghiasi2020SimpleCI}, X-Paste~\cite{Zhao2022XPasteRS}
        & \cmark
        & \cmark
        & \cmark
        & Limited ($\sim$1K)
        & Depends on real background
        & Sec~\ref{exp:compare_synthetic} \\
      \midrule
      \textbf{\name (Ours)}
        & --
        & \cmark
        & \cmark
        & \cmark
        & \cmark~($\sim$46K)
        & Unlimited (current 2.4M)
        & -- \\
      \bottomrule
    \end{tabular}
    \vspace{-1em}
  }
  \label{tab:qualitative_comparison}
\end{table*}

\subsection{Camera configuration augmentation}
After composing and relighting the scene, we apply camera configuration augmentation to simulate diverse camera intrinsics and viewing conditions, further decorrelating object appearance from semantic content.

\noindent\textbf{Random zoom (scaling and cropping).}
Starting from the focal length $f$ sampled during layout generation, we apply random scaling with factor $s \sim \mathcal{U}(1.0, 4.0)$ followed by random cropping to simulate camera zoom in. For a composite image $I$ of size $H \times W$, we first resize to $sH \times sW$ (modifying the focal length to $f' = s \cdot f$), then randomly crop back to $H \times W$. This operation ensures that object scale is not a reliable cue for category recognition.

\noindent\textbf{Depth-of-field blur.}
To simulate realistic depth-of-field effects controlled by aperture size, we apply selective Gaussian blur based on object depth. We randomly sample a focal plane depth $d_{\text{focal}}$ from the scene's depth distribution and an f-number $N \sim \mathcal{U}(1.4, 16)$ representing the aperture size (smaller f-numbers correspond to larger apertures and shallower depth-of-field). The blur kernel size for each object at depth $d$ is computed via the circle of confusion formula:
\begin{equation}
\sigma(d) = \frac{f^2}{N \cdot d_{\text{focal}}} \cdot \frac{|d - d_{\text{focal}}|}{d}
\end{equation}
where $f$ is the focal length sampled during layout generation. Objects near the focal plane remain sharp ($\sigma \approx 0$), while those farther away are progressively blurred. Smaller f-numbers (e.g., f/1.4) produce strong background blur mimicking portrait photography, while larger f-numbers (e.g., f/16) keep most objects in focus, simulating landscape photography.

These camera configuration augmentations, combined with our 3D geometric layout augmentation strategy, create a rich distribution of visual configurations that force models to learn robust, view-invariant representations rather than exploiting simple pictorial shortcuts.

\subsection{Generating region annotations}
We generate annotations for object detection (\od), instance segmentation (\seg), and visual grounding (\vg). For \od and \seg, we compute bounding boxes and masks by subtracting occluded pixels from each object's original mask. For \vg, we prompt QwQ-32B~\cite{qwq32b} with each object's bounding box, mask, category, and generation prompt to produce 3-6 attribute-based and spatial-based referring expressions per type, yielding at least 9 dense expressions per image.

\subsection{The \name segments and datasets}


\noindent\textbf{Synthetic object segments.} We generate 20M object segments in two groups: (1) \textbf{10M frequent-category segments} covering 1.6K categories from LVIS, COCO, and ADE20K, with 200 prompts per category; (2) \textbf{10M general-category segments} covering 40K categories from LAION, GQA, and Flickr30K, with 10 prompts per category. For each prompt, we synthesize three segments using FLUX with different random seeds.

\noindent\textbf{\name datasets.}
We compose these segments into five synthetic datasets: \textbf{\fc/\gc} use frequent/general category segments (Sec.~\ref{exp:ovd}, \ref{exp:vg}); \textbf{\namelvis} uses only LVIS categories (Sec.~\ref{exp:ovs}); \textbf{\sfc/\sgc} contain single-category images with multiple instances of varied attributes for intra-class referring (Sec.~\ref{exp:intra-class}); \textbf{\mixcoco} mixes COCO segments with our synthetic segments for closed-vocabulary evaluation. We use \textbf{FC-X} to denote X images from FC (e.g., \fcfiftyk = 50K images).

\section{Experiments}

We demonstrate the effectiveness of \name from following experiments: (1) We compare \name with large-scale real datasets on open-vocabulary detection (MM-Grounding-DINO on LVIS and OdinW-35), visual grounding (MM-Grounding-DINO on RefCOCO/+/g, gRefCOCO, and DoD), and instance segmentation (APE on LVIS), and also test in low-real-data regimes (Mask2Former on COCO) to show \name's effectiveness (Secs.~\ref{exp:ovd}, \ref{exp:vg}, \ref{exp:ovs}, \ref{exp:mask2former_lowdata}). (2) We propose the intra-class referring (ICR) task and benchmark, and show that \name with its controllability can improve performance on this challenging task (Sec.~\ref{exp:intra-class}). (3) To compare with other synthetic data generation methods, we evaluate \name against copy-paste, diffusion-based, and segment-based approaches (Sec.~\ref{exp:compare_synthetic}). (4) We provide fine-grained ablation studies on layout strategies, generative harmonization, and synthetic segment quality (Sec.~\ref{exp:mask2former_ablation_10k}).

We first provide a qualitative comparison of \name with existing synthetic data generation paradigms in Table~\ref{tab:qualitative_comparison}, highlighting the key advantages of our object composition approach.



\subsection{Task 1: Open-vocabulary object detection}\label{exp:ovd}

\begin{table}[t]
  \caption{\textbf{(Section~\ref{exp:ovd}) Comparison of applying different data for MM-Grounding-DINO on open-vocabulary detection tasks LVIS and ODINW-35.} Key findings: a small \name dataset (50K) yields larger LVIS AP gains than 20M GRIT and matches 200K V3Det;  scaling \name to 400K provides additional gains in LVIS (31.4) and ODINW-35 (22.8). \name complements the existing dataset as combining \name with V3Det and GRIT still brings more gain.}
  \vspace{-1em}
  \centering
  \resizebox{\columnwidth}{!}{
    \begin{tabular}{%
      l   
      c   
      cccc   
      c     
    }
    \toprule
    \multirow{2}{*}{\textbf{Data}}
      & \multirow{2}{*}{\textbf{Data Scale}}
      & \multicolumn{4}{c}{\textbf{LVIS}}
      & \textbf{OdinW-35} \\
    \cmidrule(l){3-6}\cmidrule(l){7-7}
      & 
      & AP & AP$_{\text{rare}}$ & AP$_{\text{common}}$ & AP$_{\text{frequent}}$
      & AP$_{\text{avg}}$ \\
    \midrule
    O365+GoldG (Baseline)
      & 1.4M
      & 20.1      & 10.1      & 15.3      & 29.9
      & 20.3 \\
    \midrule
    \multicolumn{7}{c}{Adding human (V3Det) or model (GRIT) annotated datasets.} \\
    \midrule
    +GRIT
      & +20M
      & 27.1  & 17.3  & 22.6  & 36.4
      & \textbf{22.8} \\
    +V3Det
      & +200K
      & \textbf{30.6} & \textbf{21.5} & \textbf{25.5} & \textbf{40.2}
      & 21.4 \\
    \midrule
    \multicolumn{7}{c}{Adding various scale SOC datasets.} \\
    \midrule
    +\fc-50K
      & +50K
      & 29.8  & 23.5 & 26.9 & 35.9
      & 20.5 \\
    +\fc-50K+\gc-50K
      & +100K
      & 30.8 & 25.6 & 27.5 & 36.7
      & 21.0 \\
    +\fc-100K
      & +100K
      & 31.0 & 26.3 & 27.8 & \textbf{36.8}
      & 21.0 \\
    +\fc-200K + \gc-200K
      & +400K
      & \textbf{31.4} & \textbf{27.9} & \textbf{28.5} & 36.1
      & 21.2 \\
    +\fc-400K
      & +400K
      & 31.0 & 25.4 & 28.2 & 36.5
      & \textbf{22.8} \\
    \midrule
    \multicolumn{7}{c}{Combining human (V3Det) and model (GRIT) annotated datasets with SOC datasets.} \\
    \midrule
       O365+GoldG+GRIT+V3Det
      & 21.6M
      & 31.9 & 23.6 & 27.6 & \textbf{40.5}
      & \textbf{23.2} \\
    +\fc-50K+\gc-50K
      & +100K
      & \textbf{33.2} & \textbf{29.8} & \textbf{30.0} & 38.3
      & 23.1 \\
    \bottomrule
    \end{tabular}
  }
  \label{tab:pivoted_with_odinw_summary}
\end{table}

\begin{table*}[!t]
    \caption{\textbf{(Section~\ref{exp:vg}) Comparison of applying different data for MM-Grounding-DINO on visual grounding benchmarks gRefCOCO, DoD (FULL), and RefCOCO/+/g avg.} Key findings: Existing datasets only yield marginal performance due to a lack of high-quality referring expression data. \name yields significantly larger gains on gRefCOCO, DoD, and RefCOCO avg compared with 20M GRIT and 200K V3Det. It is also complementary to the existing dataset.}
  \vspace{-1em}
  \centering
  \begin{adjustbox}{max width=0.95\textwidth}
    \begin{tabular}{%
      l   
      c   
      cc  
      ccccc 
      ccc  
    }
    \toprule
    \multirow{2}{*}{\textbf{Data}}
      & \multirow{2}{*}{\textbf{Data Scale}}
      & \multicolumn{2}{c}{\textbf{gRefCOCO}}
      & \multicolumn{5}{c}{\textbf{DoD (FULL)}}
      & \multicolumn{3}{c}{\textbf{RefCOCO avg.}} \\
    \cmidrule(l){3-4}\cmidrule(l){5-9}\cmidrule(l){10-12}
      &
      & P@1 (F$_1$=1, IoU$0.5$) & N$_\mathrm{acc}$
      & mAP & mAP$_\mathrm{short}$ & mAP$_\mathrm{mid}$ & mAP$_\mathrm{long}$ & mAP$_\mathrm{very\_long}$
      & P@1 & P@5 & P@10 \\
    \midrule
    O365+GoldG (Baseline)
      & 1.4M
      & 39.8 & 89.3
      & 15.6 & 17.3 & 16.7 & 14.3 & 13.1
      & 54.3 & 89.3 & 94.6 \\
    \midrule
    \multicolumn{12}{c}{Adding human (GRIT) or model (V3Det) annotated datasets.} \\
    \midrule
    +GRIT
      & +20M
      & \textbf{40.7} & \textbf{89.3}
      & \textbf{17.0} & \textbf{17.7} & \textbf{18.0} & \textbf{15.7} & \textbf{15.7}
      & \textbf{56.5} & \textbf{90.5} & \textbf{95.4} \\
    +V3Det
      & +200K
      & 40.3 & \textbf{89.3}
      & 16.7 & \textbf{17.7} & \textbf{18.0} & \textbf{15.7} & \textbf{15.7}
      & 55.0 & 89.4 & 94.6 \\
    \midrule
    \multicolumn{12}{c}{Adding various‐scale SOC datasets.} \\
    \midrule
    +\fc-50K
      & +50K
      & 41.2 & 93.9
      & 16.6 & 18.3 & 17.1 & 15.7 & 14.5
      & 54.3 & 90.3 & \textbf{95.1} \\
    +\fc-100K
      & +100K
      & \textbf{41.3} & \textbf{97.7}
      & \textbf{19.4} & \textbf{22.0} & \textbf{20.4} & \textbf{17.8} & \textbf{16.9}
      & 54.0 & \textbf{90.4} & 94.9 \\
    +\fc-50K+\gc-50K
      & +100K
      & 40.9 & 93.0
      & 17.3 & 20.0 & 18.1 & 15.7 & 14.8
      & \textbf{55.0} & 89.9 & 94.9 \\
    \midrule
    \multicolumn{12}{c}{Combining human, model‐annotated, and SOC datasets.} \\
    \midrule
    O365+GoldG+GRIT+V3Det
      & 21.6M
      & \textbf{41.0} & 89.3
      & 17.5 & 23.4 & 18.3 & 14.7 & 13.8
      & \textbf{56.4} & \textbf{90.9} & \textbf{95.8} \\
    +GRIT+V3Det+\fc-50K+\gc-50K
      & +100K
      & 40.9 & \textbf{94.4}
      & \textbf{18.9} & \textbf{23.7} & \textbf{20.5} & \textbf{15.9} & \textbf{14.3}
      & 55.1 & 89.8 & 94.8 \\
    \bottomrule
    \end{tabular}
  \end{adjustbox}
  \label{tab:pivoted_grounding_summary}
  \vspace{-1em}
\end{table*}

To study the efficacy of \name data on open-vocabulary object detection, we use MM-Grounding-DINO~\citep{zhao2024open}, a strong open‐vocabulary detector and visual grounding model that handles various types of textual input and provides bounding boxes.
We initialize the model with real-data pretrained weights and continue training on our synthetic dataset. 
We report performance on two benchmarks: (1). \textbf{LVIS}~\citep{gupta2019lvisdatasetlargevocabulary} for open‐vocabulary detection, reporting AP on the LVIS 1.0 Full Val split (including AP$_\mathrm{rare}$, AP$_\mathrm{common}$, and AP$_\mathrm{frequent}$); (2). \textbf{OdinW-35}~\citep{li2021grounded,li2022elevater} for scene‐specific detection, evaluating across 35 diverse scenarios and reporting the average AP.

\noindent \textbf{Baselines.}
We compare with 4 baseline training datasets: (1). \textbf{Object365+GoldG}: trains on Object365~\citep{9009553} (600K images, 365 open‐vocabulary categories) together with GoldG~\citep{li2022groundedlanguageimagepretraining} (800\,K vision–grounding examples mixing GQA~\citep{hudson2018gqa} and Flickr30K Entities~\citep{plummer2016flickr30kentitiescollectingregiontophrase}); (2). \textbf{Object365+GoldG+GRIT}: adds GRIT~\citep{schuhmann2022laion5bopenlargescaledataset}, a 20M example visual grounding dataset curated from LAION-2B~\cite{ziyiLaion-2B} and COYO-700M~\cite{kakaobrain2022coyo-700m}; (3). \textbf{Object365+GoldG+V3Det}: augments Object365+GoldG with V3Det~\citep{wang2023v3detvastvocabularyvisual}, a 200k human‐annotated detection dataset covering 13K categories; and (4). \textbf{Object365+GoldG+V3Det+GRIT}: combines all four real datasets~\citep{gupta2022gritgeneralrobustimage}. 

\noindent \textbf{Small amount of \name efficiently brings strong gain.}
Even at a small scale, \name delivers substantial improvements over the Object365+GoldG baseline. With only 50K synthetic examples, \name boosts LVIS AP from 20.1 to 29.8 (+9.7), and rare‐class AP from 10.1 to 23.5 (+13.4), far exceeding the +7.0 gain of GRIT (20M real examples) and matching—or even surpassing—the +10.5 gain of V3Det (200K real data). These results demonstrate that a relatively small, targeted synthetic dataset can yield gains comparable to orders of magnitude more real data, especially in the crucial rare and common categories where real‐data coverage is limited. On OdinW-35, \name-50K also improves average AP by 0.2 points and increases the number of scenarios exceeding the baseline in 20 out of 35 cases, confirming its effectiveness even in scene‐specific detection settings.

\noindent \textbf{Scaling up \name data leads to better performance.}  
Doubling \name from 50K to 100K synthetic examples yields a clear uplift on LVIS—overall AP climbs from 29.8 to 31.0 (+1.2) with rare-class AP rising from 23.5 to 26.3 (+2.8)—matching the improvement achieved by adding a 200K real-data V3Det split (+10.5 on LVIS AP). Scaling \name further to 400K synthetic examples pushes overall AP to 31.4 (+1.6 over 100K) with rare-class AP rising to 27.9 (+1.6), and increases the OdinW-35 mean AP from 21.0 at 100K to 22.8 at 400K (+1.8). These results demonstrate that \name scales efficiently in low-annotation regimes for LVIS (rare category) and continues to generalize across diverse, scene-specific contexts as its size grows.

\noindent \textbf{\name is complementary to the real datasets.}
\name complements large‐scale real datasets rather than simply duplicating their benefits. When combined with GRIT and V3Det, adding 100K synthetic examples further raises LVIS AP from 31.9 to 33.2 (+1.3) and rare‐class AP from 23.6 to 29.8 (+6.2). Crucially, this synthetic augmentation yields consistent gains on OdinW-35 as well—improving the average AP by 2.8 points over the Object365+GoldG baseline even in the presence of both GRIT and V3Det. These additive effects indicate that \name introduces novel vocabulary and contextual variations not captured by existing real datasets, thereby broadening the model’s coverage and robustness.


\subsection{Task 2: Visual grounding}\label{exp:vg}
\begin{figure*}[!t]
  \centering
    \begin{minipage}[h]{0.32\textwidth}
      \centering
      \includegraphics[width=\linewidth]{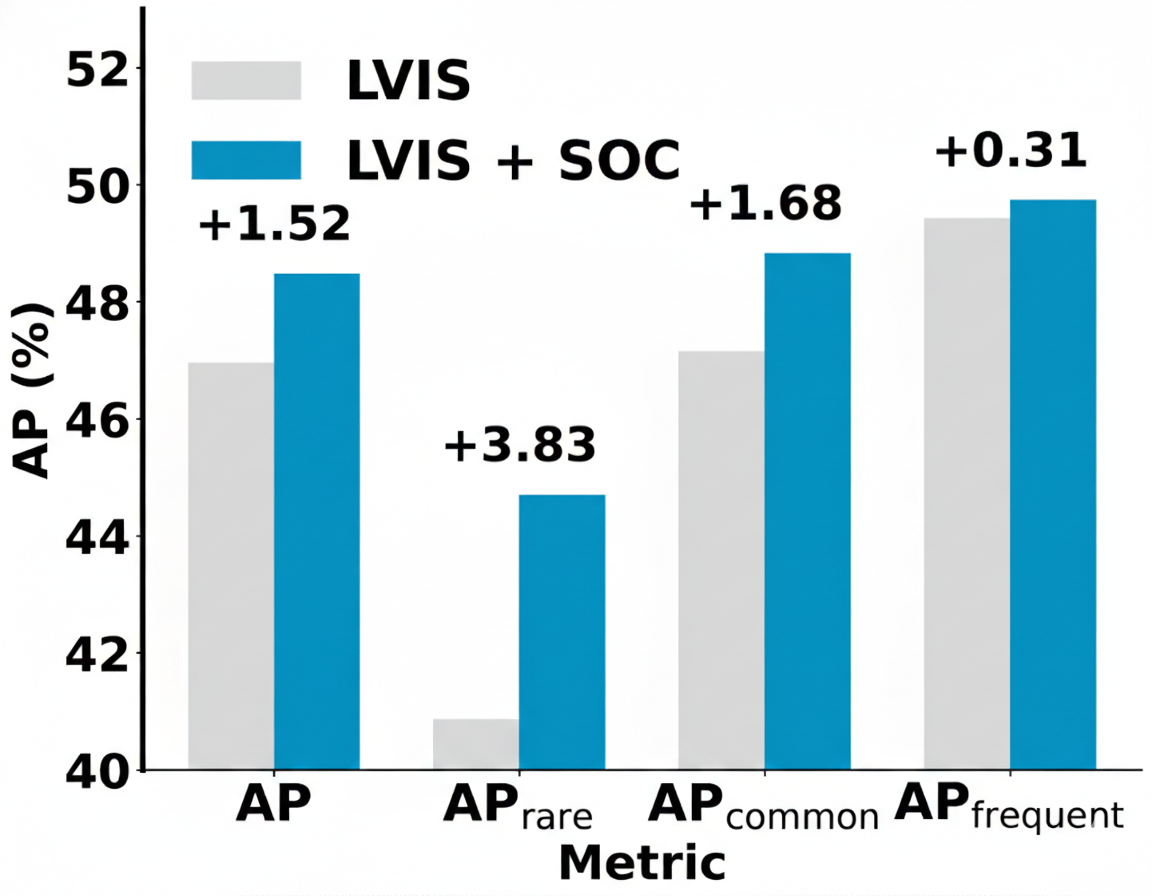}\\[-0.5ex]
      \captionof{figure}{\textbf{(Sec.~\ref{exp:ovs})} SOC consistently improves LVIS instance segmentation, with the largest gains on rare categories.}
      \label{fig:lvis-ape}
    \end{minipage}\hfill
    \begin{minipage}[h]{0.32\textwidth}
      \centering
      \includegraphics[width=\linewidth]{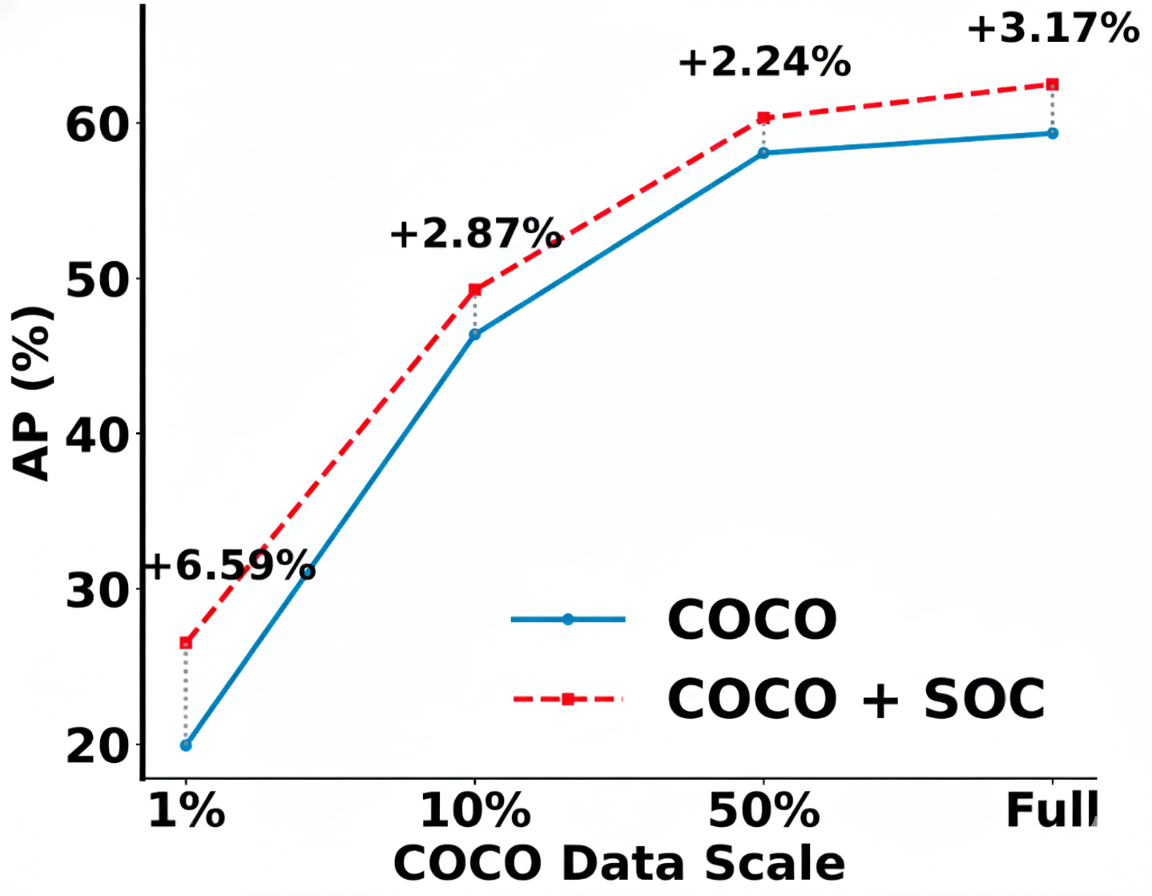}\\[-0.5ex]
      \captionof{figure}{\textbf{(Sec.~\ref{exp:mask2former_lowdata})} Combining \name synthetic segments with real COCO segments increases AP over all COCO data scales.}
      \label{fig:mask2former-coco}
    \end{minipage}\hfill
    \begin{minipage}[t!]{0.32\textwidth}
      \centering
      \vspace{-10pt}
      \resizebox{0.9\linewidth}{!}{
\begin{tabular}{lc}
  \toprule
  \textbf{Pipeline} & \textbf{AP} \\
  \midrule
  \multicolumn{2}{c}{\itshape Ablating scene layout (w/o camera aug)} \\
  \midrule
  Random 2D layout                       &  9.07           \\
  COCO layout                            &  8.60           \\
  LayoutGPT                              &  8.82           \\
  3D geometric layout augmentation       & \textbf{10.03} \textcolor{gray}{\scriptsize (+16.6\%)}   \\
  \midrule
  \multicolumn{2}{c}{\itshape Ablating camera configuration augmentation} \\
  \midrule
  w/o camera configuration augmentation  & 10.03           \\
  w/ camera configuration augmentation   & \textbf{10.58} \textcolor{gray}{\scriptsize (+5.5\%)}   \\
  \midrule
  \multicolumn{2}{c}{\itshape Ablating generative harmonization} \\
  \midrule
  w/o generative harmonization           &  6.28           \\
  w/ inpainting and relighting           & 10.58           \\
  w/ inpainting and relighting \& blending & \textbf{12.79} \textcolor{gray}{\scriptsize (+103.7\%)}  \\
  \midrule
  \multicolumn{2}{c}{\itshape Ablating object segments} \\
  \midrule
  Real segments only                     &  7.03           \\
  Real + \name~synthetic segments        & \textbf{12.79} \textcolor{gray}{\scriptsize (+81.9\%)}  \\
  \bottomrule
\end{tabular}}\\[-0.5ex]
      \captionof{table}{\textbf{(Sec.~\ref{exp:mask2former_ablation_10k})} Ablation on zero-shot COCO instance segmentation AP.}
      \label{tab:ablation_ap}
    \end{minipage}
    \vspace{-1em}
\end{figure*}

To study the visual grounding task, we use MM-Grounding-DINO and the same baseline and model training recipe as Sec.~\ref{exp:ovd}. We report results on the RefCOCO, RefCOCO+ and RefCOCOg~\cite{kazemzadeh-etal-2014-referitgame, Yu2016ModelingCI} validation splits using Precision@K (K=1,5,10); on the gRefCOCO~\cite{GREC, GRES} validation split using Precision@($F_1=1$, IoU$\geq$0.5) and no-target accuracy (how often the model correctly recognizes that there is no object to refer to); and for DoD~\cite{xie2023describedobjectdetectionliberating} benchmark, we use mAP across all lengths of description (short, mid, long, very long).
Throughout the paper, we treat Referring Expression Comprehension (REC) and Visual Grounding (VG) as equivalent and use the terms interchangeably.

\noindent \textbf{Existing large detection and grounding datasets yield only marginal improvements.}
Large detection datasets like V3Det and massive model-generated grounding datasets generated from caption pools like GRIT offer either precise labels or sheer scale, but neither delivers the targeted, precise, sentence and phrase‐level supervision needed for reliable referring‐expression grounding. Adding V3Det to Object365+GoldG yields only +0.5 P@1 on gRefCOCO (a newer dataset, distinct from RefCOCOg) and no improvement in no‐target accuracy, while GRIT’s model-generated caption-box pairs boost DoD FULL mAP by just +1.4 despite 400$\times$ more examples.

\noindent \textbf{\name provides diverse, high‐quality referring expressions that yield strong gains.}  
\name generates precise referring pairs by inferring from object attributes and spatial relationships from ground truth annotations generated through the image composing process without any human label. It immediately improves gRefCOCO no-target accuracy by +4.6  and DoD FULL mAP by +1.0. Scaling to 100K (\fc-100K) further raises no-target accuracy by +8.4 and DoD FULL mAP by +3.8. These gains per synthetic example far outperform those from GRIT (20M) and V3Det (200K), demonstrating that automatically synthesized, accurate expressions can exceed real data in both efficiency and quality. Moreover, our pipeline synthesizes expressions across different types of balance:  attribute-based (“red round ball”), spatial (“to the left of the tree”), and mix-type (“red object to the right of the child”). This targeted diversity ensures broad linguistic coverage and unlocks consistently high grounding performance across multiple benchmarks.

\subsection{Task 3: Instance segmentation}\label{exp:ovs}



We study mask-based supervision with \name for the instance segmentation task. We adopt APE~\citep{shen2023aligningpromptinguniversalvisual}, a state-of-the-art open-vocabulary segmentation and detection model pre-trained on LVIS. Since APE already sees LVIS during pre-training, this setup lets us evaluate whether our synthetic \name masks still boost performance. Our fine-tuning protocol has two stages: (1) train APE on 50K synthetic \name images that cover the same LVIS categories; (2) continue training on the LVIS v1 train split to close the domain gap. We compare this two-stage approach to a baseline trained solely on LVIS under the same FLOPs budget. All models are evaluated on the LVIS v1 validation set using overall AP, as well as AP$_\text{rare}$, AP$_\text{common}$, and AP$_\text{frequent}$.

\noindent \textbf{\name continuously improve the model's performance on LVIS.} As could be seen from Figure~\ref{fig:lvis-ape},
incorporating 50K synthetic \name images yields a particularly large gain on rare categories: AP$_\text{rare}$ increases from 40.87 to 44.70 (+3.83), while overall AP rises from 46.96 to 48.48 (+1.52) and AP$_\text{frequent}$ has a +0.31 improvement. This pattern occurs because synthetic data can be generated to cover underrepresented classes, thereby mitigating the LVIS long-tail imbalance, whereas frequent classes already have ample real examples and thus benefit less from additional synthetic augmentation.

\subsection{Task 4: Small‐vocabulary, limited‐data regimes}\label{exp:mask2former_lowdata}


\noindent

Real-world segmentation applications  (e.g., surveillance cameras) only need a small set of categories and have limited annotation budgets for the dataset curation, in this section, we evaluate \name specifically in a small-vocabulary, limited-data regime to investigate its effectiveness under these practical constraints.
We adopt Mask2Former~\cite{cheng2022maskedattentionmasktransformeruniversal}, a widely-used closed-vocabulary segmentation model, for the instance segmentation task. Specifically, we compare a Mask2Former-ResNet-50~\cite{he2016deep} baseline trained solely on COCO instance-segmentation subsets across $4$ different data scales (1K 1\%), (10K (10\%), 50K (50\%), or all COCO images) against a setting augmented with our \name data. This augmented data is generated by mixing real segments collected from corresponding splits and synthetic segments from \name synthetic segments within 80 COCO categories.

\noindent\textbf{\name consistently improves models at different real data scales and extremely well on limited-data regimes.}
\name consistently improves model performance across different real-data scales and is particularly effective in limited-data regimes. As Figure~\ref{fig:mask2former-coco} shows, COCO + \name performs exceptionally well when using only 1\% of COCO data, yielding a 6.59\% gain. Moreover, the boost grows by roughly 3\% at each subsequent data scale, suggesting that our augmentation approach is effective across all scales.

\subsection{Task 5: Intra‐class referring expression, a challenging visual grounding task}\label{exp:intra-class}


\begin{figure*}[!t]
  \centering
    \begin{minipage}[h]{0.58\textwidth}
      \centering
      \includegraphics[width=\linewidth]{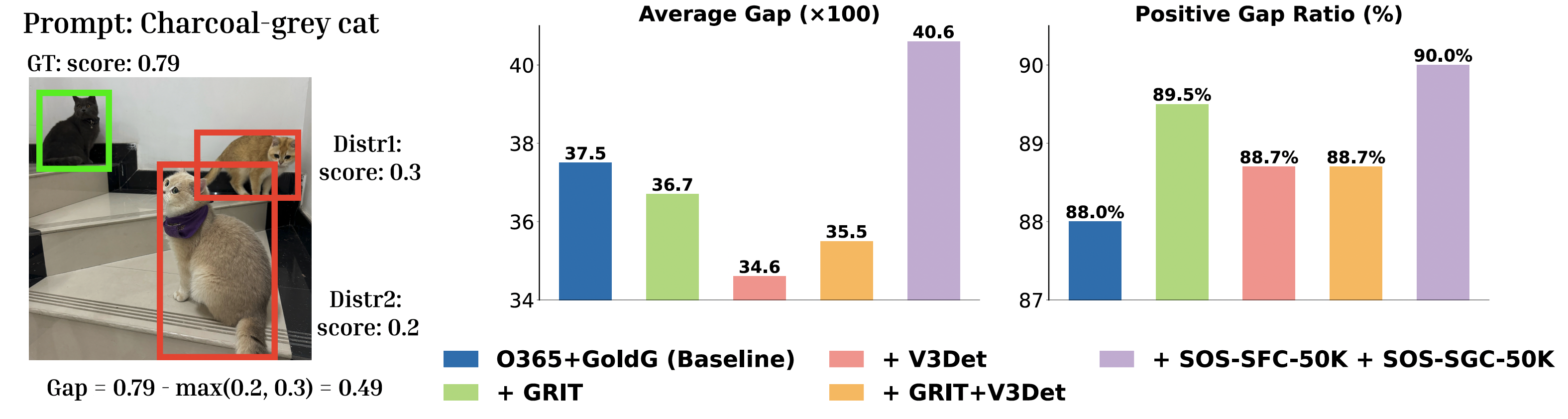}\\[-0.5ex]
      \captionof{figure}{\textbf{(Section~\ref{exp:intra-class})} Intra-class referring demands fine-grained attribute discrimination among same-category instances (e.g., picking out the "charcoal-grey cat"). Augmenting with SOC synthetic pairs (+\fc-50K+\gc-50K) yields the largest improvements in both metrics.}
      \label{fig:intra-class}
    \end{minipage}\hfill
    \begin{minipage}[h]{0.37\textwidth}
      \centering
      \vspace{0pt}
      \resizebox{\linewidth}{!}{    \begin{tabular}{lccc}
      \toprule
      \textbf{Method} &
      \textbf{COCO Seg} &
      \textbf{LVIS-Mini Det} &
      \textbf{gRefCOCO VG} \\
      \midrule
      Copy-Paste~\cite{Ghiasi2020SimpleCI}
        & 9.32
        & 35.2
        & -- \\
      X-Paste~\cite{Zhao2022XPasteRS}
        & 9.41
        & 37.2
        & -- \\
      SegGen~\cite{ye2024seggensuperchargingsegmentationmodels}
        & 9.73
        & 36.8
        & -- \\
      SynGround~\cite{he2024learningsyntheticdatavisual}
        & --
        & --
        & 40.1 / 89.2 \\
      \midrule
      \textbf{\name (Ours)}
        & \textbf{12.79}
        & \textbf{38.6}
        & \textbf{41.2 / 93.9} \\
      \bottomrule
    \end{tabular}

}\\[-0.5ex]
      \captionof{table}{\textbf{(Section~\ref{exp:compare_synthetic})} Quantitative comparison with synthetic baselines. COCO Seg uses Mask2Former (AP), LVIS-Mini Det and gRefCOCO VG use MM-Grounding-DINO (AP for Det, P@1/N\_Acc for VG).}
      \label{tab:compare_copy_paste_xpaste}
    \end{minipage}
      \vspace{-1em}
\end{figure*}

\noindent \textbf{Intra‐class Referring Expression.}
Current visual grounding benchmarks—such as RefCOCO—can be circumvented. For example, to locate a ``green car,'' a model might ignore the attribute ``green'' and rely solely on the noun ``car.'' This shortcut breaks down when multiple cars of different colors and makes are present in the scene. We call this scenario \emph{intra-class referring}, since it requires fine-grained attribute discrimination and is therefore more challenging.
To systematically evaluate this use case,
we curate an “intra-class referring expression” benchmark from COCO and OpenImages V7~\cite{openimages} by selecting around 100 images that contain multiple instances of the same category exhibiting distinct attributes (colour, shape/pose, subtype).  Each instance is tightly boxed and manually labeled with its key attribute variation (e.g.\ “red” vs.\ “blue”, “round” vs.\ “elongated”), yielding per‐instance category labels, bounding‐box coordinates, and fine‐grained attribute annotations. 
We propose two metrics for this new task:
\textbf{Average Gap}:
For each image, we compare the model’s confidence of the correct box (IoU $\ge0.5$) against the highest-scoring same-category distractor. The gap between these two confidence scores measures how much more certain the model is about the true object than about any distractor. We report the average of this margin across all images;
\textbf{Positive Gap Ratio}:
We also calculate the percentage of images in which the ground-truth box receives the highest confidence score of all same-category candidates. This tells us in how many cases the model correctly prioritizes the intended instance over its distractors.

\noindent \textbf{\name can generate data that targetly addresses intra‐class referring errors.}
As shown in Figure \ref{fig:intra-class}, adding large-scale auxiliary data (GRIT, V3Det, or both) yields a negligible or negative impact on AvgGap (down to 34.6–36.7) and only slight Positive Gap Ratio gains. By contrast, fine‐tuning on 100 K targeted \sfc-50K+\sgc-50K samples raises AvgGap by 3.1 points to 40.6. It boosts the Positive Gap Ratio to 90\%, proving that synthetic data tailored to intra‐class attribute is essential for improvement.

\subsection{Comparison with other synthetic pipelines}\label{exp:compare_synthetic}

As shown in Table~\ref{tab:compare_copy_paste_xpaste} and Table~\ref{tab:qualitative_comparison}, \name substantially outperforms copy-paste baselines: on COCO instance segmentation (Mask2Former), \name achieves 12.79 AP vs. 9.3 AP for Copy-Paste and 9.4 AP for X-Paste; on LVIS-Mini open-vocabulary detection (MM-Grounding-DINO), \name reaches 38.6 AP vs. 35.2 AP for Copy-Paste and 37.2 AP for X-Paste.

\noindent\textbf{Key advantages.} \name uniquely combines: (1) \textbf{Open-vocabulary coverage} (40K+ categories vs. $\sim$1.3K for copy-paste methods), (2) \textbf{Pixel-accurate annotations} (unlike diffusion methods where generated images deviate from input masks), (3) \textbf{Multi-object composition} with realistic lighting (vs. single-object focus in Subject200K), and (4) \textbf{Multi-task support} (detection, segmentation, and grounding vs. single-task diffusion methods).

\subsection{Ablation study}\label{exp:mask2former_ablation_10k}

We conduct an ablation study on COCO instance segmentation using Mask2Former trained only on 10K \name data from scratch (Table~\ref{tab:ablation_ap}), ablating four key components: (1) scene layout strategies, (2) camera configuration augmentation, (3) generative harmonization, and (4) synthetic object segment quality. Our 3D geometric layout augmentation (10.03 AP) outperforms random 2D layout (9.07), COCO layout (8.60), and LayoutGPT (8.82). Adding camera configuration augmentation improves to 10.58 AP (+5.5\%). Generative harmonization yields the largest gain: without it, AP is 6.28; with inpainting and relighting, 10.58; adding blending reaches 12.79 (+103.7\%). Our synthetic segments (12.79 AP) substantially outperform real segments alone (7.03 AP, +81.9\%). We use the optimal configuration for all subsequent experiments. More analysis in the Appendix.

\section{Conclusion}

We presented \name, a scalable pipeline that turns synthetic object segments into richly annotated images for detection, instance segmentation, and visual grounding.  By composing segments with 3D geometric layout augmentation and camera configuration augmentation, generative harmonization with mask-area-weighted blending, \name yields pixel-perfect masks, boxes, and diverse referring expressions—without human labeling.  We show through experiments that across three common tasks (open-vocabulary detection, instance segmentation, referring expression) a small slice of \name data consistently outperforms or complements much larger real or synthetic image corpora. We additionally show that \name is effective in a limited-data regime and can help a diagnostic visual grounding task with targeted data synthesis.

\appendix

\clearpage
\begin{center}
{\LARGE\bfseries Appendix}
\end{center}
\vspace{1em}

\setcounter{page}{1}

\section{Limitations and Future Work}  

\paragraph{Synthetic–real domain gap.}
Our relighting and blending strategies bring synthetic images closer to real photographs than directly pasting segments onto images. The remaining differences in surface texture, material response, and global illumination are subtle. These minor artifacts have a limited impact on downstream tasks compared to the performance boost we observed relative to real data, and they can be further reduced in future work by integrating state-of-the-art diffusion models.

\paragraph{3D Coherence.}
Although we operate on standalone 2D object segments and directly paste them into images without explicit 3D constraints, this simplification poses little difficulty for current 2D-focused benchmarks (e.g., detection, segmentation, referring expressions). For applications requiring full 3D coherence, such as depth estimation or novel-view synthesis, future extensions could incorporate 3D geometry priors or 3D assets into the generation pipeline.

\paragraph{Object Interactions.}
For tasks such as detection, grounding, and segmentation, inter-object relations are not a primary domain feature. Therefore, for simplicity, our pipeline currently treats objects independently and does not model inter-object relations. In practice, this has a negligible impact on object-centric recognition tasks, and future work could address this by adding fine-grained relation control—using diffusion models in our pipeline—if we wish to extend it to relation recognition.

\paragraph{Panoptic segmentation.}  
In this work, we focus on object-level recognition. We produce high-quality object segments for discrete “things,” but do not yet annotate amorphous “stuff” regions or enforce an all-pixel partitioning, as we are aiming to improve on object detection, instance segmentation, and visual grounding. This limitation does not impede most instance-level or semantic benchmarks, and can be readily addressed by integrating off-the-shelf “stuff” predictors or by extending our annotation pipeline to include full-scene panoptic labels in future releases.

\section{Detailed Related work}

\noindent \textbf{Open-Vocabulary detection and segmentation.}
Traditional object detection~\cite{girshick2015fast,ren2016faster} and segmentation~\cite{he2017mask, cheng2022maskedattentionmasktransformeruniversal} methods typically target predefined categories from standard benchmarks (e.g., COCO~\cite{lin2015microsoftcococommonobjects}, OpenImages~\cite{Kuznetsova_2020}, Object365~\cite{9009553}, LVIS~\cite{gupta2019lvisdatasetlargevocabulary}). Transformer-based architectures, such as DETR~\cite{Carion2020EndtoEndOD} and its derivatives~\cite{Zhu2020DeformableDD, Zhang2022DINODW, Lv2023DETRsBY,Huang2024DQDETRDW}, have revolutionized end-to-end detection. Recent advances in vision-language models trained via contrastive language supervision~\cite{radford2021learning, Jia2021ScalingUV} have expanded detection to open-vocabulary settings~\cite{Zareian2020OpenVocabularyOD}. Approaches include distillation of pretrained vision-language knowledge into region-level representations~\cite{Gu2021OpenvocabularyOD,Zhong2021RegionCLIPRL}, adapting contrastive models for detection~\cite{Minderer2022SimpleOO}, integrating language-conditioned queries into DETR framework~\cite{Zang2022OpenVocabularyDW}, and leveraging abundant image-level annotations to overcome detection annotation scarcity~\cite{Zhou2022DetectingTC,Lin2022LearningOA}.
Building upon these approaches, scaling pre-training datasets has significantly advanced open-vocabulary models~\cite{li2021grounded, Yao2022DetCLIPDV, Yao2023DetCLIPv2SO}. GLIP~\cite{li2021grounded} combines detection with grounded language-image pretraining, introducing the ODInW benchmark~\cite{li2021grounded,li2022elevater}. 
Datasets like V3Det~\cite{wang2023v3detvastvocabularyvisual}, with over 13,000 categories, further push open-vocabulary capabilities. MM-Grounding-DINO~\cite{Liu2023GroundingDM, zhao2024open}, a combination of DINO~\cite{Zhang2022DINODW} and grounded pretraining, represents state-of-the-art performance alongside real-time solutions like Yolo-World~\cite{Cheng2024YOLOWorldRO}. Similarly, open-vocabulary segmentation (OVS) models~\cite{Zhou2021ExtractFD, ghiasi2022scalingopenvocabularyimagesegmentation, Li2022LanguagedrivenSS, Liang2022OpenVocabularySS} now incorporate language supervision, evaluated on standard semantic segmentation datasets (COCO-Stuff~\cite{Caesar2016COCOStuffTA}, ADE20K~\cite{8100027,zhou2018semanticunderstandingscenesade20k}, PASCAL VOC~\cite{everingham2011pascal}, PASCAL Context~\cite{Mottaghi2014TheRO}, and Cityscapes~\cite{Cordts2016Cityscapes}). Universal models like APE~\cite{shen2023aligningpromptinguniversalvisual} unify grounding, segmentation, and detection under diverse prompts~\cite{Zou2022GeneralizedDF, Zou2023SegmentEE}.

\noindent \textbf{Referring expression and Grounding.}
Referring expression comprehension involves localizing image objects using natural language descriptions~\cite{Yu2016ModelingCI}. Early datasets, such as ReferItGame~\cite{kazemzadeh-etal-2014-referitgame}, evolved into standard benchmarks like RefCOCO, RefCOCO+, and RefCOCOg~\cite{Yu2016ModelingCI, Mao2015GenerationAC}, annotated over COCO images. Flickr30K Entities~\cite{plummer2016flickr30kentitiescollectingregiontophrase} and Visual Genome~\cite{Krishna2016VisualGC} facilitate visual grounding with detailed region annotations. Earlier methods employed modular decomposition~\cite{Hu2016ModelingRI} (e.g., subject and relationships detected individually in MattNet~\cite{Yu2018MAttNetMA}), but transformer-based models now dominate~\cite{Su2019VLBERTPO, li2019visualbertsimpleperformantbaseline, lu2019vilbertpretrainingtaskagnosticvisiolinguistic, chen2020uniteruniversalimagetextrepresentation, deng2022transvgendtoendvisualgrounding, Kamath2021MDETRM}. MDETR introduced a modulated detector conditioned directly on textual queries~\cite{Kamath2021MDETRM}. Recent open-vocabulary models, including GLIP and MM-Grounding-DINO, have also been adapted to this task~\cite{Huang2024DQDETRDW, li2021grounded, Liu2023GroundingDM, zhao2024open, deng2022transvgendtoendvisualgrounding, du2022visualgroundingtransformers}. Concurrently, the introduction of datasets like GoldG~\cite{li2022groundedlanguageimagepretraining} provides essential human annotations for grounding tasks. Multimodal large language models (MLLMs) have achieved state-of-the-art REC performance by directly generating bounding box coordinates from text~\cite{Wu2022GRiTAG, You2023FerretRA, Ma2024GromaLV}. Recent efforts like KOSMOS-2~\cite{Peng2023Kosmos2GM} leverage synthetic data to overcome annotation scarcity, although verifying synthetic annotations remains challenging~\cite{Yao2024DetCLIPv3TV}.

\noindent \textbf{Synthetic datasets for detection, segmentation, and visual grounding.} 
Synthetic datasets have emerged to provide large-scale annotations without manual labeling, with early examples such as SYNTHIA~\cite{Ros2016TheSD} and GTA5~\cite{Richter2016PlayingData} offering useful segmentation labels via game-engine rendering but revealed domain gaps when applied to real-world images. Virtual KITTI 2~\cite{Cabon2020VKITTI2} and Synscapes~\cite{Wrenninge2018Synscapes} significantly improved realism for driving scenes through photorealistic and physically-based rendering, and Hypersim~\cite{Roberts_2021_ICCV} offered extensive indoor data with dense segmentation, depth, and normal annotations. More recently, diffusion-based synthesis has produced high-resolution, photorealistic images with controllable layouts~\cite{Rombach2021HighResolutionIS, Nichol2021GLIDETP, Ramesh2022HierarchicalTI, Li2023GLIGENOG}, realistic illumination variations~\cite{Kocsis2024LightItIM, zhang2025scaling}, and scalability to open-vocabulary and pseudo-labeled data~\cite{He2022IsSD, Azizi2023SyntheticDF, lee2025molmoactactionreasoningmodels,
Fan2023ScalingLO, zheng2025brokentokenslanguagemodel,
Zhang2023AddingCC, Jiang2022PseudoQGP, Peng2023Kosmos2GM, du2025paintoutsideboxsynthesizing, gao2024generatesceneevaluatingimproving, zhang2024provision}.

Early efforts at synthetic data relied heavily on 3D simulators, which require painstakingly crafted assets and still struggle to capture the diversity of real-world scenes. For purely 2D approaches, one branch resorts to simple copy-and-paste augmentations, transplanting segmented objects into real images \cite{Dvornik2018ModelingVC, Fang2019InstaBoostBI, Ghiasi2020SimpleCI}. However, such methods remain constrained by the quality of available segments and produce conspicuous seams around pasted regions, causing models to exploit edge cues rather than true object semantics. X-Paste~\cite{Zhao2022XPasteRS} improves segment integrity and diversity by using diffusion models to enrich object segments before pasting, but it still randomly places segments and therefore does not fully resolve realism or seam-artifact issues. Moreover, because these methods are defined as augmentation pipelines applied to existing images, they are limited by the scale of the real data and restricted to a narrow set of categories (e.g., only 1.3 K categories on LVIS and COCO).

A complementary line of work generates entire scenes with diffusion models or directly using real images, but applies off-the-shelf detectors to harvest pseudo-labels for bounding boxes. Wang et al.~\cite{Wang2024LearningVG} automatically produce referring annotations using VLMs. He et al.~\cite{he2024learningsyntheticdatavisual} leverage diffusion models to create varied images and use detectors to generate bounding-box labels. While these approaches achieve higher image fidelity, their pseudo-labels are noisy and yield smaller performance gains compared to real-image annotations, and also limit them to only high-level grounding tasks, instead of dense segmentation and detection tasks.

In contrast, our framework generates over 20 million high-quality object segments across more than 46K categories, each annotated with fine-grained metadata. We integrate generative harmonization (IC-Light for background inpainting and global relighting) and mask-area-weighted blending to boost photorealism and to provide precise masks, bounding boxes, and referring expressions. This strategy combines the advantages of both copy-paste and diffusion-based approaches while avoiding unrealistic artifacts and inaccurate labels. It not only overcomes the scale and category limitations of prior augmentation methods but also delivers data that is fully suitable for open-vocabulary detection, segmentation, and visual grounding in authentic real-world scenes.

\section{Ablation of Scaling Only Object Segments or Images}

In our main experiment, since we generate 20M object segments, we simultaneously scale the number of object segments and the number of images—each object segment is used only once within the composed images—and observe that larger datasets yield better performance. To disentangle the individual effects of scaling images versus object segments, we conduct two ablation studies:


\begin{figure*}[tbp]
  \centering
  \includegraphics[width=\linewidth]{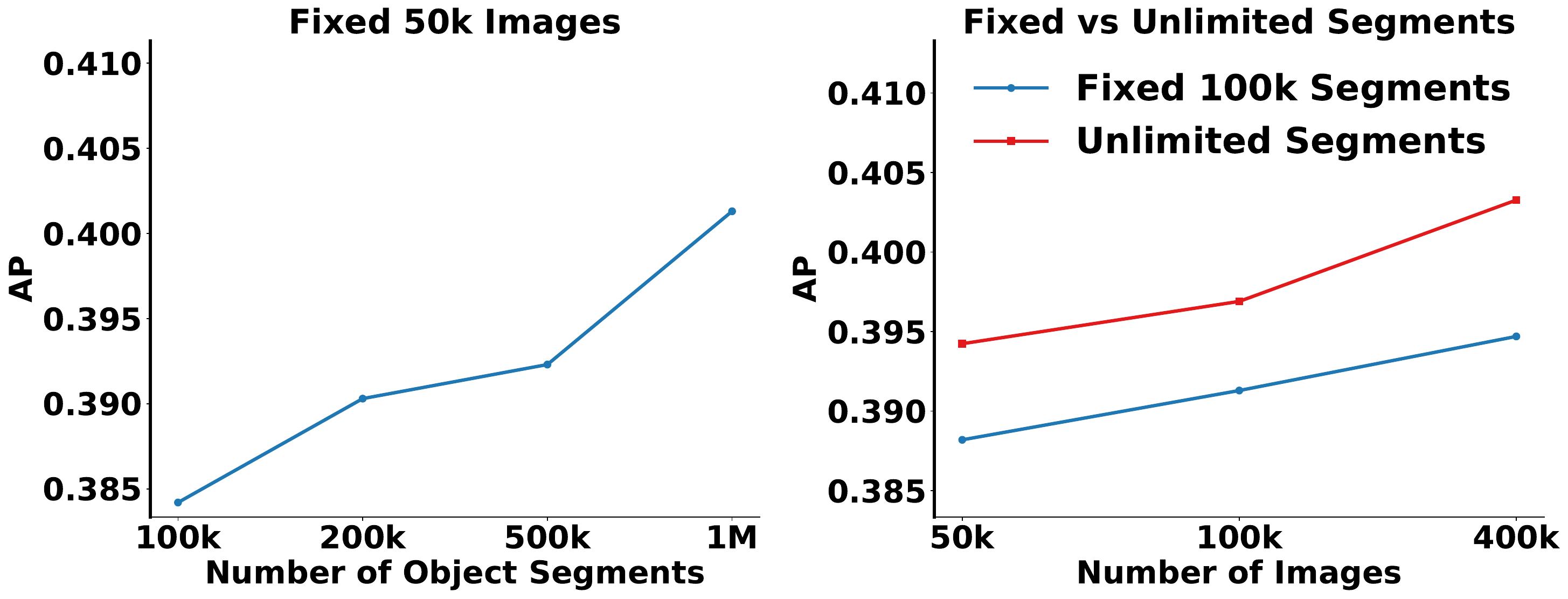}
  \caption{Impact of scaling object segments (left) and images (right) on AP. In the left plot, we fix the number of images to 50\,K and vary segments from 100\,K to 1\,M; in the right plot, we compare fixed–100\,K segments versus an unlimited‐segment regime as we scale images from 50\,K to 400\,K. Models are evaluated on \textbf{LVIS-Mini Val} and report AP. }
  \label{fig:ablation_scaling}
\end{figure*}

\paragraph{Fixing the number of images and scaling object segments.}  
We fix the image count at 50\,K and vary the total number of object segments through four settings: 100\,K, 200\,K, 500\,K, and 1\,M. In each case, we sample 20 segments per image. Consequently, when only 100\,K segments are available, each segment must be reused $20 \times 50$\,K $/$ 100\,K $=$ 10 times; with 200\,K segments the reuse factor falls to 5 times; at 500\,K it drops to 2 times; and at 1\,M segments every sampled segment is unique. As shown in Figure~\ref{fig:ablation_scaling} (left), AP increases monotonically from 0.3842 at 100\,K segments to 0.4013 at 1\,M segments, with the largest incremental gain occurring between 500\,K and 1\,M segments. This pattern indicates diminishing returns beyond 500\,K but still underscores the benefit of richer segment diversity.

\paragraph{Fixing the number of segments and scaling images.}
Conversely, we fix the segment count at 100\,K and scale the number of images through 50\,K, 100\,K, 200\,K, and 400\,K, mirroring the unlimited‐segment setup in our main experiments. Under this fixed–100\,K–segment regime (Figure~\ref{fig:ablation_scaling}, right, blue curve), AP climbs modestly from 0.3882 to 0.3947 as images quadruple, reflecting the limited benefit of reusing the same segments. In contrast, when we allow an unlimited pool of segments (red curve), AP grows more strongly from 0.3942 to 0.4033 over the same image range. The larger gap under the unlimited‐segment condition demonstrates that adding fresh segments is crucial to fully leverage additional images.

\paragraph{Analysis.}  
These ablation studies reveal that, while scaling images alone yields modest gains when segment diversity is capped, increasing the number of unique object segments provides larger improvements in AP. The strongest performance arises when both image count and segment diversity are scaled together, confirming the advantage of our joint scaling strategy for open‐vocabulary detection and segmentation.

\section{Compare with Other Synthetic Method}
We position \name within the broader synthetic landscape through qualitative comparisons in Table~\ref{tab:segment_comparison} and Table~\ref{tab:pipeline_comparison}.

\subsection{Comparison with Synthetic Object Segments}
\begin{table*}[t]
  \caption{Qualitative comparison of synthetic \emph{object‑segment} libraries.  
  \name offers orders‑of‑magnitude more segments, explicit category control, and multi‑view diversity—key to the photorealistic, richly annotated composites described in §\,4.1.}
  \centering
  \begin{adjustbox}{max width=\textwidth}
    \begin{tabular}{lcccc}
      \toprule
      \textbf{Library} &
      \textbf{Diffusion model} &
      \textbf{Category} &
      \textbf{Scale (segments)} &
      \textbf{Viewpoint diversity}\\
      \midrule
      Subject 200K           & FLUX & Less than 1000 & 200K  & \xmark\\
      \midrule
      \textbf{\name (ours)}    & FLUX & 46K+ & 20M   & \cmark\\
      \bottomrule
    \end{tabular}
  \end{adjustbox}
  \label{tab:segment_comparison}
\end{table*}

Subjects 200K \cite{tan2024omini} contains roughly 200K FLUX‑generated segments, but they cover <1K categories and are captured from only a handful of canonical viewpoints. Because the library offers neither broad category coverage nor multi‑view variation, it limits the semantic and geometric diversity attainable when composing new scenes.

By contrast, \name delivers \textbf{20M} segments spanning \textbf{46K+} categories and explicitly samples multiple camera angles for every object prompt, ensuring rich viewpoint diversity. This two‑orders‑of‑magnitude increase in scale—coupled with fine‑grained control over category and viewpoint—unlocks far more varied, photorealistic composites than any prior segment library.

\paragraph{Quantitative comparison with Subject200K.}
To validate the quality advantage of our synthetic object segments, we conduct a controlled experiment on COCO instance segmentation using Mask2Former trained on 10K synthetic images. As shown in Table~\ref{tab:subject200k_comparison}, when combining real segments with synthetic object segments, \name achieves 12.79 AP, outperforming Subject200K (12.06 AP) by +0.73 AP (+6.1\% relative improvement). This demonstrates that our object segment pipeline—with its broader category coverage (46K+ vs. <1K) and multi-view diversity—produces higher-quality segments that lead to better downstream performance.

\begin{table}[h]
  \centering
  \caption{Quantitative comparison with Subject200K on COCO instance segmentation (Mask2Former, zero-shot). All models are trained on 10K synthetic images composed from real segments + synthetic object segments.}
  \label{tab:subject200k_comparison}
  \begin{tabular}{lcccc}
    \toprule
    \textbf{Object Segments} & \textbf{AP} & \textbf{AP$_{\text{S}}$} & \textbf{AP$_{\text{M}}$} & \textbf{AP$_{\text{L}}$} \\
    \midrule
    Real segments only & 7.03 & 0.55 & 3.15 & 7.22 \\
    Real + Subject200K & 12.06 & 1.43 & 6.53 & 14.57 \\
    Real + \name (ours) & \textbf{12.79} & \textbf{1.58} & \textbf{7.71} & \textbf{15.74} \\
    \midrule
    \textbf{Improvement} & \textbf{+0.73} & \textbf{+0.15} & \textbf{+1.18} & \textbf{+1.17} \\
    \bottomrule
  \end{tabular}
\end{table}

\subsection{Comparison with Synthetic Data Pipelines}

\begin{table*}[h]
  \caption{Qualitative comparison of complete \emph{synthetic‑data pipelines}.  
  \name is the only approach that simultaneously offers fine‑grained control, open‑vocabulary coverage, realistic integration (relighting + blending), and pixel‑accurate annotations across \emph{all} dense‑vision tasks.}
  \centering
  \begin{adjustbox}{max width=\textwidth}
    \begin{tabular}{lcccccl}
      \toprule
      \textbf{Pipeline} &
      \textbf{Method} &
      \textbf{Fine‑grained control} &
      \textbf{Open‑vocabulary} &
      \textbf{Realistic integration} &
      \textbf{Accurate masks/boxes} &
      \textbf{Supported task(s)}\\
      \midrule
      Simulator‑rendered (e.g.\ SYNTHIA) & Game‑engine scenes & \cmark & \xmark & \cmark & \cmark & DET, SEG\\
      Simple Copy‑Paste      & Paste onto real images & \cmark & \xmark & \xmark & \cmark & DET, SEG\\
      X‑Paste                & Paste onto real images & \cmark & \xmark & \xmark & \cmark & DET, SEG\\
      SynGround              & Diffusion images       & \xmark & \cmark & \cmark & \xmark & VG\\
      Learning VG            & Diffusion images       & \xmark & \cmark & \cmark & \xmark & VG\\
      \midrule
      \textbf{\name (ours)}    & Composing new images    & \cmark & \cmark & \cmark & \cmark & DET, SEG, VG\\
      \bottomrule
    \end{tabular}
  \end{adjustbox}
  \label{tab:pipeline_comparison}
\end{table*}

\textbf{Simulator‑rendered datasets}—SYNTHIA~\cite{Ros2016TheSD}, GTA5~\cite{Richter2016PlayingData}, Virtual KITTI 2~\cite{Cabon2020VKITTI2}, Synscapes~\cite{Wrenninge2018Synscapes}, and the indoor‑focused Hypersim~\cite{Roberts_2021_ICCV}—deliver pixel‑accurate masks and boxes by design, yet they remain confined to pre‑built domains (mostly driving or indoor scenes) and a closed object vocabulary; a noticeable photo‑to‑sim gap still emerges when models are applied to real photographs.  

\textbf{Copy‑paste families} such as Simple Copy‑Paste~\cite{Ghiasi2020SimpleCI}, InstaBoost~\cite{Fang2019InstaBoostBI}, and the diffusion‑refined X‑Paste~\cite{Zhao2022XPasteRS} transplant real segments into new images.  They are inexpensive and controllable, but the pasted objects often betray seam artefacts or lighting clashes.  Because the pipeline augments a fixed real corpus, its scale is capped by the number of host images and its category list rarely exceeds \textasciitilde1.3 K classes from LVIS/COCO.  

\textbf{Diffusion‑plus‑pseudo‑label pipelines} invert the recipe.  Methods like SynGround~\cite{he2024learningsyntheticdatavisual} and Learning VG~\cite{Wang2024LearningVG} first synthesise entire scenes with text‑to‑image models~\cite{Rombach2021HighResolutionIS, Nichol2021GLIDETP, Ramesh2022HierarchicalTI}, then harvest boxes or phrases via detectors or VLMs.  Although the images are photorealistic and open‑vocabulary, the labels inherit detector noise and therefore suit only coarse grounding rather than dense segmentation.

\textbf{\name} is best seen as an \emph{object‑centric composition pipeline}. The detailed comparisons are shown in Table~\ref{tab:segment_comparison} and Table~\ref{tab:pipeline_comparison}.

The outcome is a dataset that matches simulators in annotation fidelity, rivals diffusion images in photorealism, and, through open‑vocabulary, layout‑controlled synthesis, vastly outstrips prior copy‑paste methods in diversity and scale. These qualities underpin the improvements reported in Sec.~\ref{exp:ovd}--\ref{exp:intra-class}, where \name demonstrates strong performance against both synthetic-based methods and real-image training data (e.g., GRIT, V3Det)—a stark contrast to existing synthetic datasets that are commonly treated as mere complements to real data.

\subsection{Visual Realism: FID Score Analysis}

To quantify the visual realism of our synthetic images, we computed Fréchet Inception Distance (FID)~\cite{Heusel2017GANsTB} scores on a 1K-sample subset comparing \name against Copy-Paste and X-Paste baselines. As shown in Table~\ref{tab:fid_comparison}, \name achieves an FID of 131.93, substantially lower than both Simple Copy-Paste (165.55) and X-Paste (166.03). This indicates that even without explicitly optimizing for photorealism, \name produces images that are more distributionally aligned with natural image statistics than alternative synthetic pipelines.

\begin{table}[h]
  \centering
  \caption{FID scores (lower is better) comparing visual realism of synthetic data generation methods. Computed on 1K samples against real image distribution.}
  \begin{tabular}{lc}
    \toprule
    \textbf{Method} & \textbf{FID Score} \\
    \midrule
    Simple Copy-Paste & 165.55 \\
    X-Paste & 166.03 \\
    \midrule
    \textbf{\name (Ours)} & \textbf{131.93} \\
    \bottomrule
  \end{tabular}
  \label{tab:fid_comparison}
\end{table}

We note that \name optimizes for annotation accuracy (pixel-perfect masks), compositional diversity (46K+ categories), and training effectiveness rather than photorealism alone. The strong downstream performance against both synthetic-based and non-synthetic-based methods (Sec.~\ref{exp:ovd}) validates this design choice.

\section{Details of \name Pipeline}

\subsection{Details of Mask-Area-Weighted Blending Algorithm}
To achieve realistic lighting, we adopt IC‑Light~\citep{zhang2025scaling}, a foreground‑conditioned diffusion model that ingests an image containing foreground objects, generates a matching background with a text prompt, and relights the composite to achieve photorealism.

However, IC-Light introduces two challenges in multi-object scenes: (1) strong relighting may distort fine details of small objects, making them unrecognizable, and (2) excessive relighting can alter object colors, breaking consistency with color-based descriptions from the original data.

To address this, we use a segment‐area‐aware blending process: We introduce a blending weight \(\alpha_i\in[0,1]\) for each object mask \(M_i\), which controls the degree of relighting applied to that object. Smaller objects receive higher \(\alpha_i\), preserving more of their original appearance; larger objects receive lower \(\alpha_i\), allowing more of the relit image to show through.

\begin{enumerate}[leftmargin=*,itemsep=0.25em,topsep=0.25em]
  \item \textsc{Size-based weighting.}
    \begin{align*}
      r_i &= \frac{\operatorname{area}(M_i)-A_{\min}}{A_{\max}-A_{\min}}, && r_i\in[0,1], \\
      \alpha_i &= \alpha_{\min}
      + (\alpha_{\max}-\alpha_{\min})\,
        \sigma\!\left(s\,(r_i-\tfrac12)\right).
    \end{align*}
    Smaller objects (low $r_i$) get higher $\alpha_i$ (milder relighting). The sigmoid groups “small” segments together instead of scaling linearly by size.

  \item \textsc{Lab-space blend.}
    For $p\in M_i$, convert
    \(
      I_O(p),I_R(p)\to (L_O,a_O,b_O),(L_R,a_R,b_R).
    \)
    Blend lightness and chroma separately:
    \begin{align*}
      L_{\text{out}} &= \alpha_i L_O + (1-\alpha_i) L_R, \\
      \mathbf{c}_{\text{out}} &=
      (1-\beta_i)\,\mathbf{c}_O + \beta_i\,\mathbf{c}_R,
      \quad \mathbf{c}=(a,b),\ \ \beta_i\ll 1,
    \end{align*}
    then convert $(L_{\text{out}},\mathbf{c}_{\text{out}})$ back to RGB.

  \item \textsc{Background passthrough.}
    \(
      I_{\text{out}}(p)=I_R(p)
      \ \text{for}\ p\notin\bigcup_i M_i.
    \)
\end{enumerate}

This segment-area‑aware blending ensures that small segments avoid over‑relighting by IC-Light (preserving detail) while larger ones receive stronger adjustments, and by blending only the luminance (with a small chroma factor \(\beta_i\)) in CIELAB space, we maintain perfect color fidelity and small object details, but also increase the photorealism.

\subsection{Details of 3D Geometric Layout Augmentation}

\paragraph{Robustness-first design: Why photorealistic layouts are not our goal.}
A central design principle of \name is that photorealistic spatial layouts are neither necessary nor desirable for training robust vision models. Real-world photographs contain strong statistical regularities (e.g., cars appear large and near image bottoms, small objects cluster on surfaces) that models can exploit as shortcuts. Our 3D geometric layout augmentation deliberately breaks these correlations by sampling object depth, size, and position independently of category, creating compositions that may appear ``unnatural'' but force models to learn view-invariant, category-robust representations.

Generative harmonization plays a complementary role: it eliminates low-level copy-paste artifacts (lighting inconsistencies, boundary discontinuities) that would provide trivial cues for distinguishing synthetic data, without constraining spatial layouts to match photographic distributions. This separation is intentional—harmonization ensures sufficient visual coherence to avoid shortcut learning from obvious artifacts, while our layout strategy ensures sufficient diversity to avoid shortcut learning from statistical priors.

The effectiveness of this approach is demonstrated by our results (Sec.~\ref{exp:ovd}--\ref{exp:intra-class}), where \name-trained models outperform both synthetic-based methods and models trained on real images (GRIT, V3Det). These results suggest that photorealistic layouts may actually be detrimental to robustness, as they reintroduce the pictorial biases that limit generalization.

\paragraph{3D scene modeling with category-independent sampling.}
Our 3D geometric layout augmentation strategy models each composite image as a 3D scene where depth and spatial position are sampled independently of object category. This ensures objects of the same category appear at diverse depths, sizes, and positions, preventing category-specific pictorial patterns. For each image, we sample 5-20 object segments (matching COCO/SA-1B distributions) using balanced category sampling to avoid bias.

Each object category $c$ has a commonsense physical size range (e.g., cars: 4-5m, cups: 10-20cm) generated by Qwen2.5-32B. The complete pipeline is:

\begin{enumerate}[leftmargin=*,itemsep=0.25em,topsep=0.25em]
  \item \textbf{Sample camera focal length}: $f \sim \mathcal{U}(f_{\min}, f_{\max})$
  \item \textbf{Define maximum depth}: $D_{\max} = \alpha \cdot f$ (where $\alpha$ is a scaling constant in meters/pixel)
  \item \textbf{Define depth ranges} (in meters):
    \begin{itemize}
      \item Close: $[0.1 D_{\max}, 0.3 D_{\max}]$
      \item Middle: $[0.3 D_{\max}, 0.6 D_{\max}]$
      \item Far: $[0.6 D_{\max}, D_{\max}]$
    \end{itemize}
  \item \textbf{For each object segment $i$ of category $c_i$}:
    \begin{itemize}
      \item Sample physical size: $S_i \sim \mathcal{N}(\mu_{c_i}, \sigma_{c_i})$
      \item Sample depth $d_i$ from one of the three ranges, following COCO/SA-1B distribution (40\% close, 35\% middle, 25\% far)
      \item Sample 3D position: $(X_i, Y_i) \sim \mathcal{U}(X_{\min}, X_{\max}) \times \mathcal{U}(Y_{\min}, Y_{\max})$ (in meters)
      \item Project to 2D via perspective projection:
        \begin{equation*}
        x_i = f \cdot \frac{X_i}{d_i}, \quad y_i = f \cdot \frac{Y_i}{d_i}, \quad s_i = f \cdot \frac{S_i}{d_i}
        \end{equation*}
        where $(x_i, y_i)$ is the 2D center position and $s_i$ is the apparent size in pixels
    \end{itemize}
  \item \textbf{Enforce constraints}: If an object's apparent size is too small/large, or if it completely occludes another object ($\text{IoU}(M_i, M_j) \geq 0.9$), resample its 3D position and depth
\end{enumerate}

This approach ensures that object scale is determined by 3D geometry (depth + physical size) rather than category, breaking spurious correlations like ``cars appear large and near the bottom.''

\subsection{Details of Camera Configuration Augmentation}

After composing and relighting the scene, we apply camera configuration augmentation to simulate diverse camera intrinsics and viewing conditions. Each augmentation (random zoom and depth-of-field blur) is applied independently with 30\% probability.

\paragraph{Random zoom (scaling and cropping).}
Starting from the focal length $f$ sampled during layout generation, we apply random scaling with factor $s \sim \mathcal{U}(1.0, 4.0)$ followed by random cropping to simulate camera zoom in. For a composite image $I$ of size $H \times W$:
\begin{enumerate}[leftmargin=*,itemsep=0.25em,topsep=0.25em]
  \item Resize to $sH \times sW$ (modifying the focal length to $f' = s \cdot f$)
  \item Randomly crop back to $H \times W$
\end{enumerate}
This operation ensures that object scale is not a reliable cue for category recognition.

\paragraph{Depth-of-field blur.}
To simulate realistic depth-of-field effects controlled by aperture size, we apply selective Gaussian blur based on object depth:
\begin{enumerate}[leftmargin=*,itemsep=0.25em,topsep=0.25em]
  \item Generate a depth map for the composed image: We first use Depth Anything V2 to predict relative depth for the entire scene, then scale it such that the predicted depth of the farthest object matches its sampled depth $d_{\max}$ from the 3D scene modeling. Finally, we replace each object region with its exact sampled depth $d_i$. This approach ensures physical consistency with the layout while allowing natural depth variation in the background
  \item Randomly sample a focal plane depth $d_{\text{focal}}$ from the scene's depth distribution
  \item Sample an f-number $N \sim \mathcal{U}(1.4, 16)$ representing the aperture size (smaller f-numbers = larger apertures = shallower depth-of-field)
  \item Compute blur kernel size for each pixel at depth $d$ via the circle of confusion formula:
    \begin{equation*}
    \sigma(d) = \frac{f^2}{N \cdot d_{\text{focal}}} \cdot \frac{|d - d_{\text{focal}}|}{d}
    \end{equation*}
    where $f$ is the focal length sampled during layout generation and $d$ is the absolute depth in meters from the depth map
\end{enumerate}

Objects near the focal plane remain sharp ($\sigma \approx 0$), while those farther away are progressively blurred. Smaller f-numbers (e.g., f/1.4) produce strong background blur mimicking portrait photography, while larger f-numbers (e.g., f/16) keep most objects in focus, simulating landscape photography.

These camera configuration augmentations, combined with our 3D geometric layout augmentation strategy, create a rich distribution of visual configurations that force models to learn robust, view-invariant representations.

\begin{figure*}[!t]
  \centering
  \includegraphics[width=0.8\linewidth]{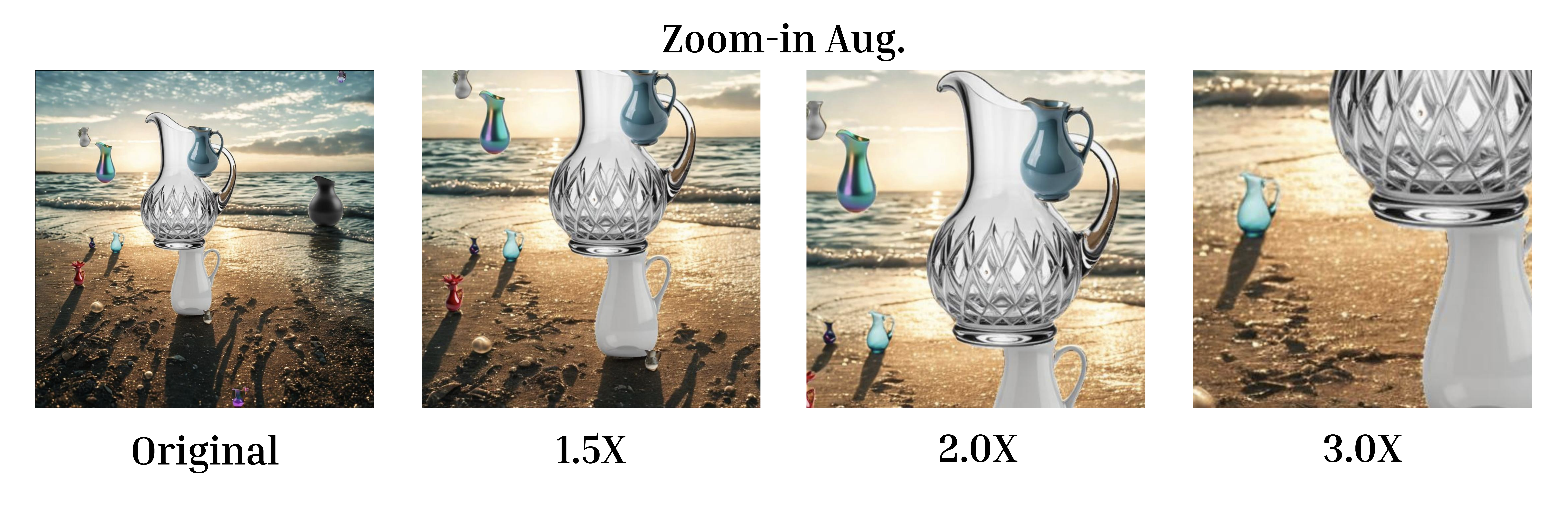}
  \caption{Examples of random zoom augmentation. Starting from the composed image (left), we apply random scaling with factor $s \sim \mathcal{U}(1.0, 4.0)$ followed by random cropping (right). This simulates camera zoom in and ensures object scale is not a reliable cue for category recognition.}
  \label{fig:zoom_aug_examples}
\end{figure*}

\begin{figure*}[!t]
  \centering
  \includegraphics[width=0.8\linewidth]{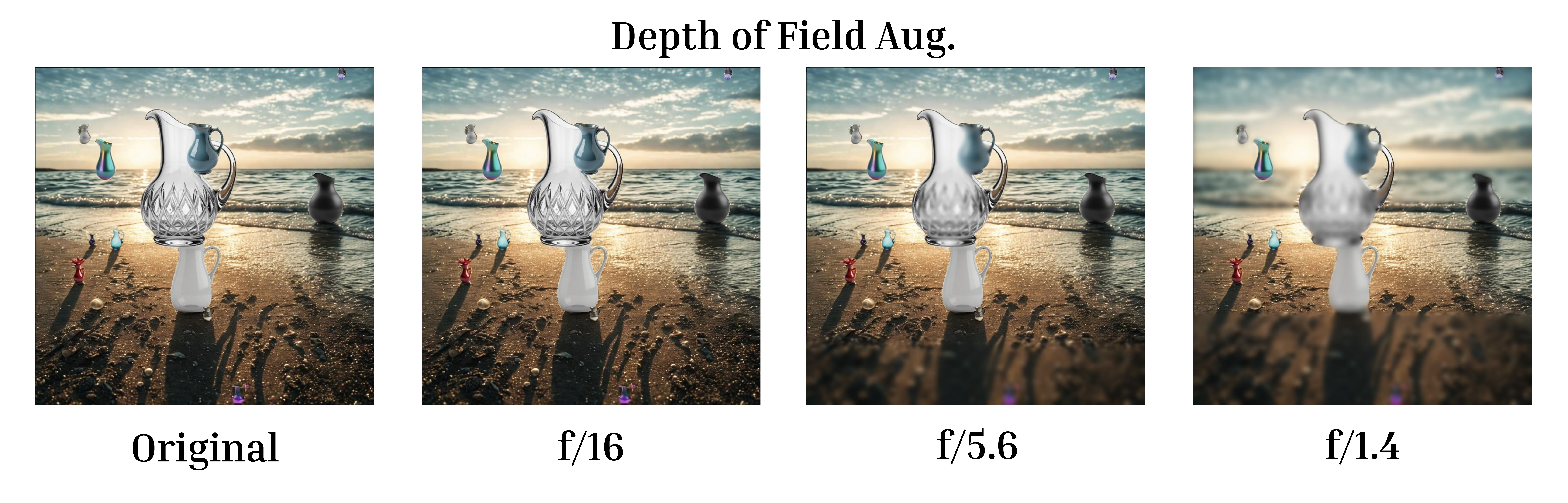}
  \caption{Examples of depth-of-field blur augmentation. Starting from the composed image (left), we apply selective Gaussian blur based on depth (right). Objects near the focal plane remain sharp while those farther away are progressively blurred, simulating realistic camera aperture effects with varying f-numbers.}
  \label{fig:dof_aug_examples}
\end{figure*}

\subsection{Prompts used in \name}
We provide our prompts used in belowing images.

\begin{figure*}[tbp]
  \centering
  \includegraphics[width=\linewidth]{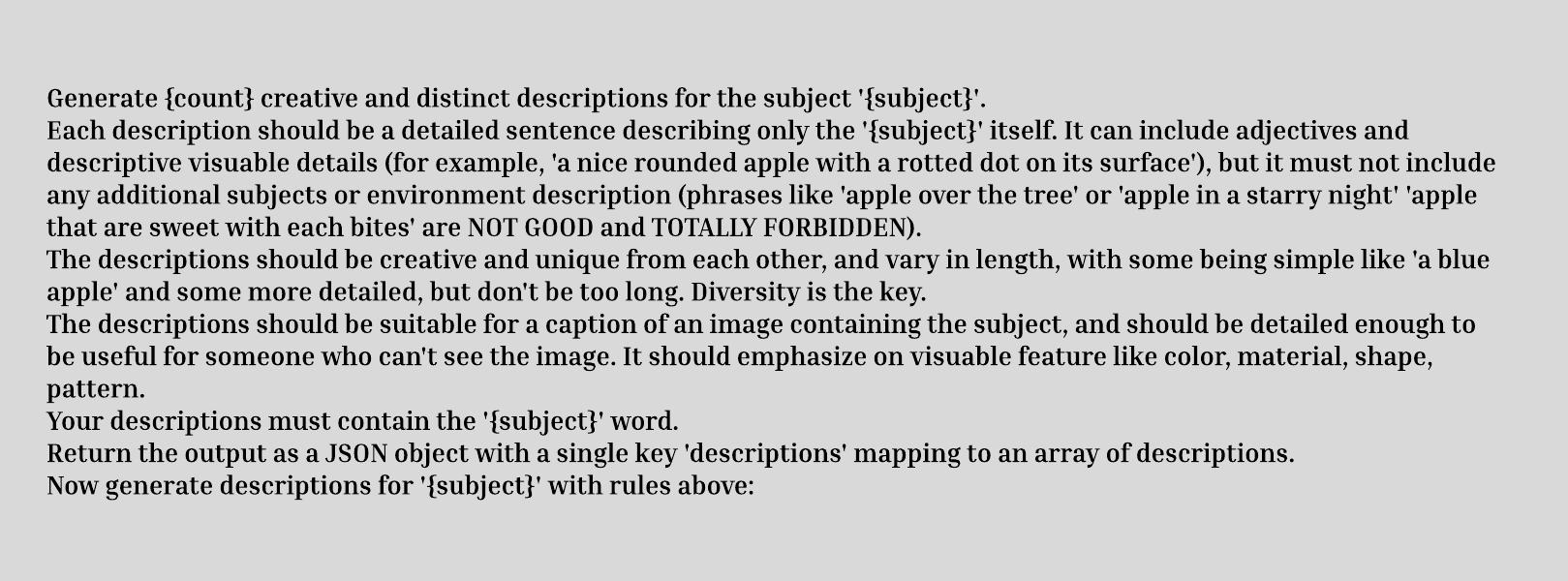}
  \caption{Prompt for generating diverse text descriptions.}
  \label{fig:prompt_generate_description}
\end{figure*}

\begin{figure*}[tbp]
  \centering
  \includegraphics[width=\linewidth]{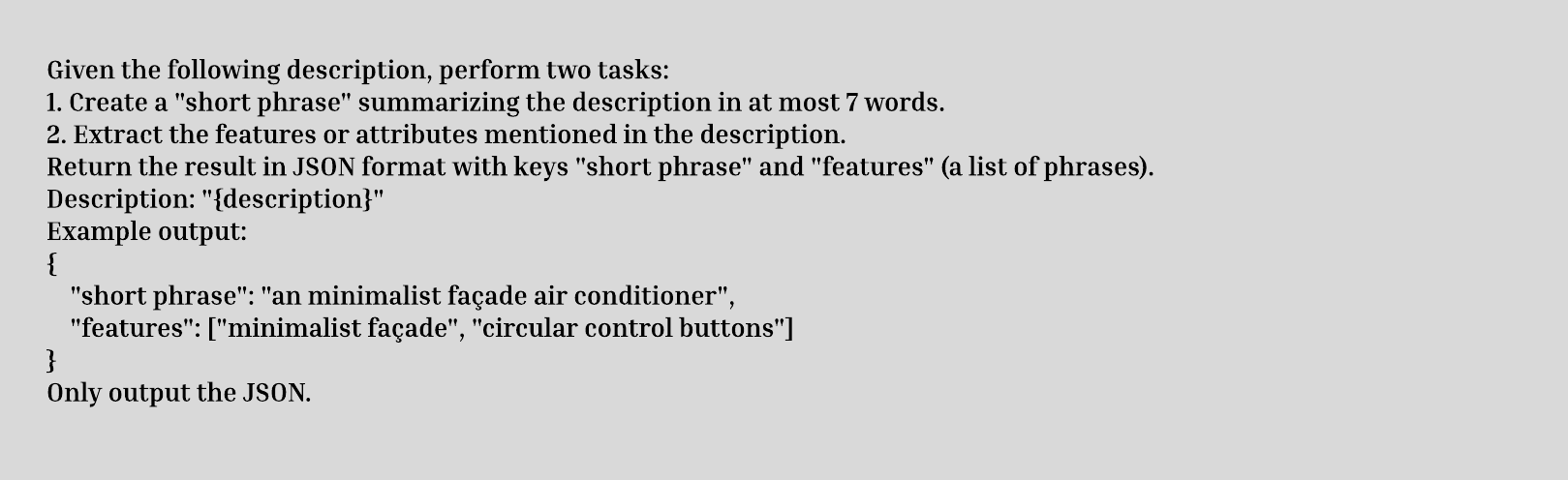}
  \caption{Prompt for extracting features and generating shortening phrases.}
  \label{fig:prompt_phrase_shortening}
\end{figure*}

\begin{figure*}[tbp]
  \centering
    \includegraphics[
      width=\textwidth,
      height=\textheight,
      keepaspectratio
    ]{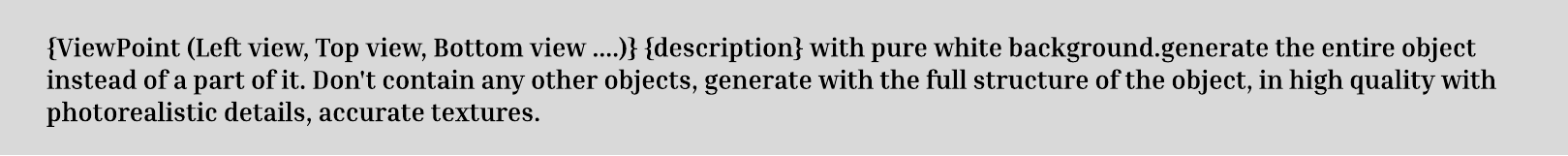}
  \caption{Prompt for generating object segments.}
  \label{fig:prompt_generate_object}
\end{figure*}

\begin{figure*}[h]
  \centering
    \includegraphics[
      width=\textwidth,
      height=\textheight,
      keepaspectratio
    ]{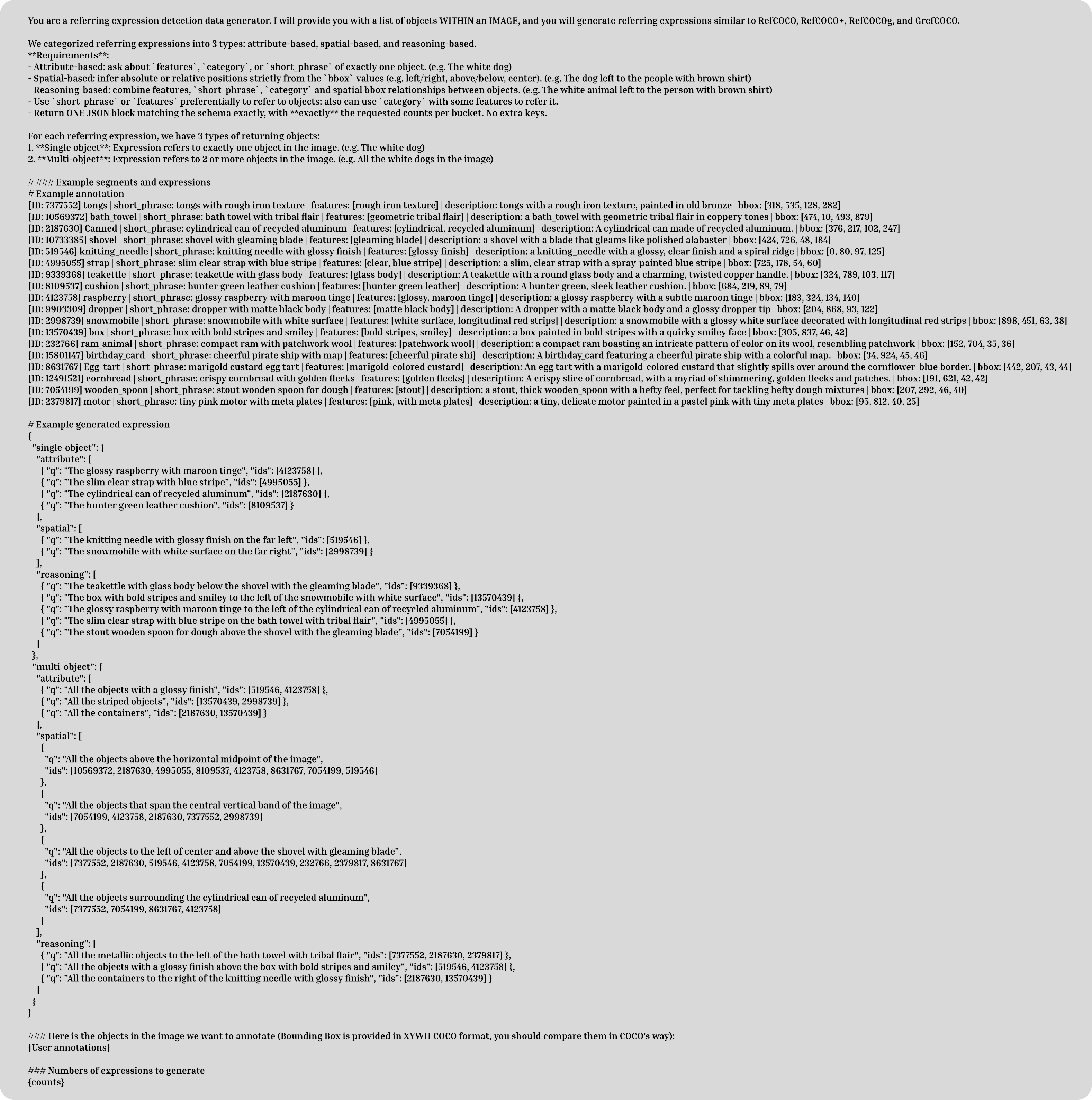}
  \caption{Prompt for generating referring expressions.}
  \label{fig:prompt_referring_expression}
\end{figure*}

\section{Details of Experiments}
In this section, we provide the training and evaluation details of the experiments conducted in the previous sections. 
\subsection{Details of Task 1: Open Vocabulary Object Detection}
\paragraph{Training details.} We use the training scripts provided by the official repo of MM-Grounding-DINO, with 8xH100 GPUs; We use the MM-Grounding-DINO-Tiny model, initialized with pretrained weights on Object365 and GoldG (with additional pretraining on O365 + GoldG + GRIT + V3Det as complementary data). During the \name synthetic-data stage, we use a batch size of 128: for each batch, 70\% of the samples are drawn from our \name dataset and 30\% from Object365 or GoldG to improve training stability. When training on both the FC and GC splits, we sample 35\% from each split to form the 70\% synthetic-data portion. Furthermore, for each synthetic bounding-box annotation, 66\% are paired with the category label (e.g., \textit{apple}) and the remaining 33\% with a short phrase (e.g., \textit{the red apple}) to diversify the training signal. This configuration is applied at the 50K, 100K, and 400K \name data scales, each run lasting 10 epochs. We use an initial learning rate of $4\times10^{-4}$, reduced to $4\times10^{-5}$ at the sixth epoch, and apply a learning-rate multiplier of 0.1 to both the visual and language backbones. Due to the syn2real gap, this stage does not directly improve benchmark performance but injects \name information into the model. Afterward, we fine-tune the model on the original Object365 and GoldG datasets for 3 epochs, sampling equally from Object365, GoldG, and our synthetic data (1:1:1). We use a learning rate of $2\times10^{-4}$, again scaling the visual and language backbone rates to 0.1 of the overall rate. After this stage, we observed strong performance gains over benchmarks. For the baseline with V3Det and GRIT, we directly use the weights provided by the MM-Grounding-DINO repo.

To validate that any performance gains stem from \name rather than extended training on real data, we also test on using only Object365 and GoldG, but matched for FLOPs and learning rates of synthetic data training, which only keeps the performance on pretrained weights, demonstrating the effectiveness of \name data.

\subsection{Details of Task 2: Visual Grounding}
\paragraph{Training details.} We follow the same setup as Task 1 for Visual Grounding, but after the original synthetic-data stages, we further train the models for five epochs on the same images, but using referring expressions as the training signal. And then follow the same real data fine-tune stage.

We also test that training with our data in Task 1 (using category and short phrase as signal instead of referring expression) doesn't improve model performance on the visual grounding benchmark. Only training on \name referring expressions brings the performance boost.

\subsection{Details of Task 3: Instance Segmentation}

\paragraph{Data generation.}
For 50K \namelvis we used. We generated only by using the synthetic object segments that are in the LVIS categories. The other data configurations are the same as the \fc and \gc.

\paragraph{Training details.} 
All experiments were conducted using the training scripts provided by the official APE repository on 4 NVIDIA H100 GPUs. We initialized this from the APE-L\_A checkpoint, which was pre-trained jointly on LVIS, COCO, Object365, OpenImages, and Visual Genome. We load those weights, set a batch size of 8, and apply a halved learning-rate schedule compared to the default (the official APE training uses batch size 16). We sample LVIS and \namelvis data equally (50\% each) and train for 30000 steps, then continue training solely on LVIS for an additional 10000 steps. For the baseline, we train directly on LVIS for 40000 steps.

\subsection{Details of Comparison with Synthetic Baselines and Ablation Study}

\paragraph{COCO instance segmentation (Mask2Former).}
For both the comparison with synthetic baselines (§4.7) and the ablation study (§4.8), we use the same experimental setup to ensure fair comparison. All models are trained on 10K synthetic images generated by each method, initialized from ImageNet-pretrained ResNet-50 weights, and trained from scratch without using any COCO images. We use the official Mask2Former training scripts on 4 A100 GPUs with batch size 32 and learning rate 0.0001. All models are evaluated zero-shot on the COCO validation set and report AP.

\paragraph{LVIS-Mini open-vocabulary detection (MM-Grounding-DINO).}
For the comparison with synthetic baselines (§4.7), we train MM-Grounding-DINO on 50K synthetic images generated by each method. We follow the same training setup as Task 1 (§4.1) and evaluate on the LVIS-Mini validation set, reporting AP.

\paragraph{Visual grounding on gRefCOCO (MM-Grounding-DINO).}
For the comparison with synthetic baselines (§4.7), we train MM-Grounding-DINO on 50K synthetic images with referring expressions. We follow the same training setup as Task 2 (§4.2) and evaluate on gRefCOCO, reporting Precision@(F$_1$=1, IoU$\geq$0.5) and no-target accuracy.

\subsection{Details of Task 4: Small-Vocabulary, Limited-Data Regimes}

\paragraph{Training details.}  We train Mask2Former (R–50 backbone) using the official Mask2Former training scripts on 4 A100 GPUs. Because the repository does not include a four‑GPU configuration, we initialize only from the ImageNet‑pretrained ResNet‑50 weights and train the remaining components from scratch. We apply the same batch size (32) and learning rate (0.0001) to both our method and the baseline, ensuring a fair comparison.

The baseline model is trained solely on COCO and is stopped when its validation AP has not improved for 2,000 consecutive iterations.

For our setup, we first perform 4,000 training iterations on real COCO images at each data scale (1K, 10K, 50K, and the full COCO set) to obtain an initial representation. Training then continues exclusively on our \textsc{cocomix} data until the training loss fails to decrease for 2,000 iterations, signalling a plateau. Finally, we fine‑tune the model on its scale of COCO data and terminate once the performance on the test set no longer improves.

\subsection{Details of Task 5: Intra-class referring}

\begin{figure*}[h]
  \centering
  \begin{subfigure}[b]{0.48\textwidth}
    \centering
    \includegraphics[width=\linewidth]{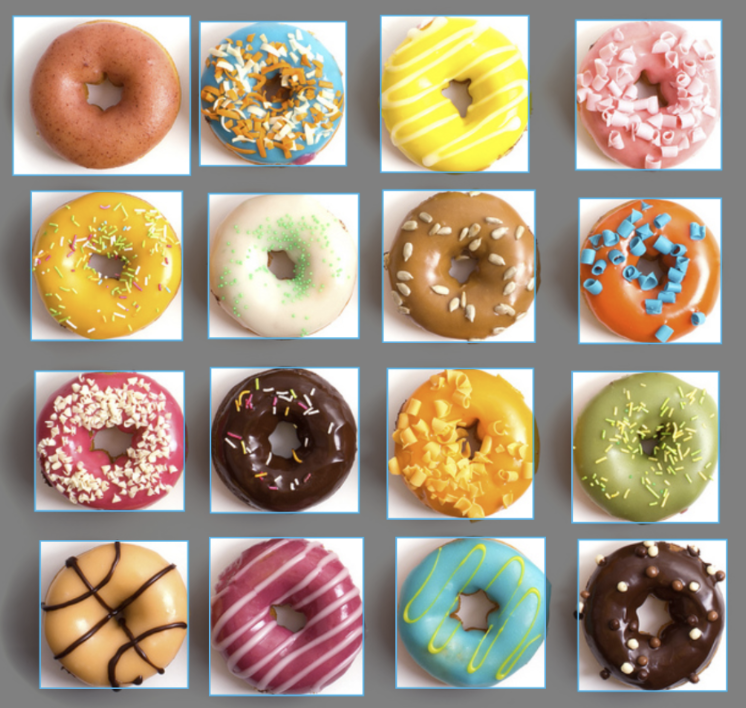}
    \caption{Multiple donuts.}
    \label{fig:prompt_referring_expression_donut}
  \end{subfigure}
  \hfill
  \begin{subfigure}[b]{0.48\textwidth}
    \centering
    \includegraphics[width=\linewidth]{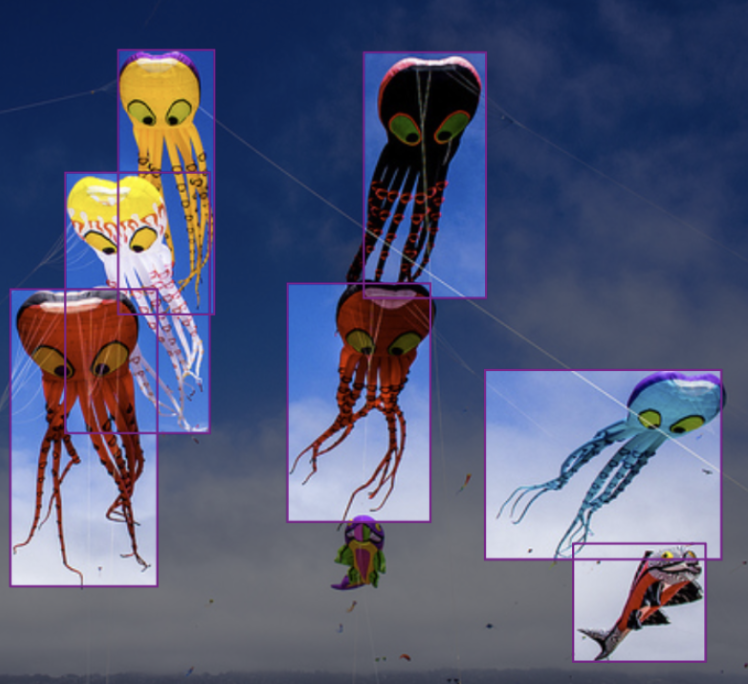}
    \caption{Multiple kites.}
    \label{fig:prompt_referring_expression_kite}
  \end{subfigure}
  \caption{Examples for the intra‑class referring benchmark.}
  \label{fig:prompt_referring_expression_row}
\end{figure*}

\paragraph{Training details.}
We follow the setup from Task 1, but instead of training over \fc and \gc, we train over \sfc and \sgc, which are images with same-category-different-attributes objects specifically generated to solve the intra-class problems. 
\begin{figure*}[!t]
  \centering
  \begin{subfigure}[b]{0.48\textwidth}
    \centering
    \includegraphics[width=\linewidth]{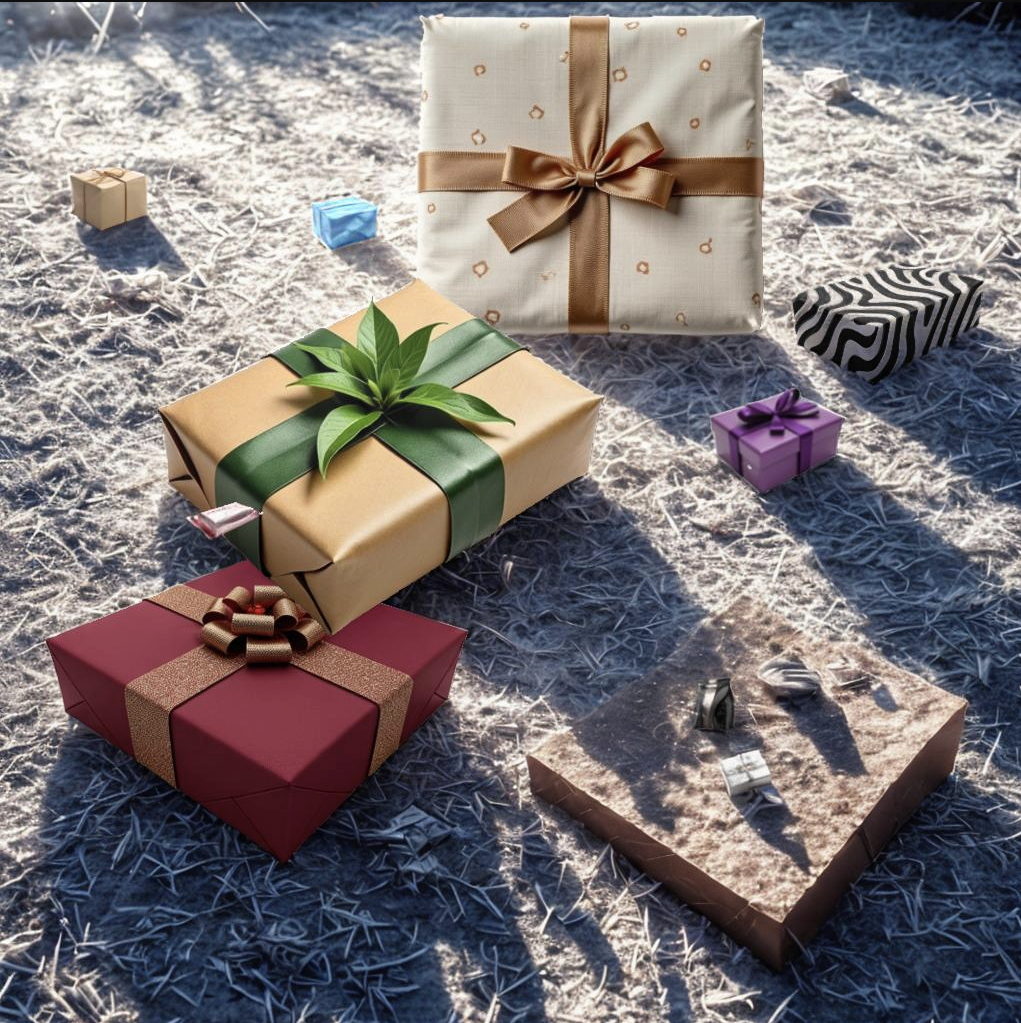}
    \caption{Traininig data sampling from \sfc and \sgc.}
    \label{fig:prompt_referring_expression_our1}
  \end{subfigure}
  \hfill
  \begin{subfigure}[b]{0.48\textwidth}
    \centering
    \includegraphics[width=\linewidth]{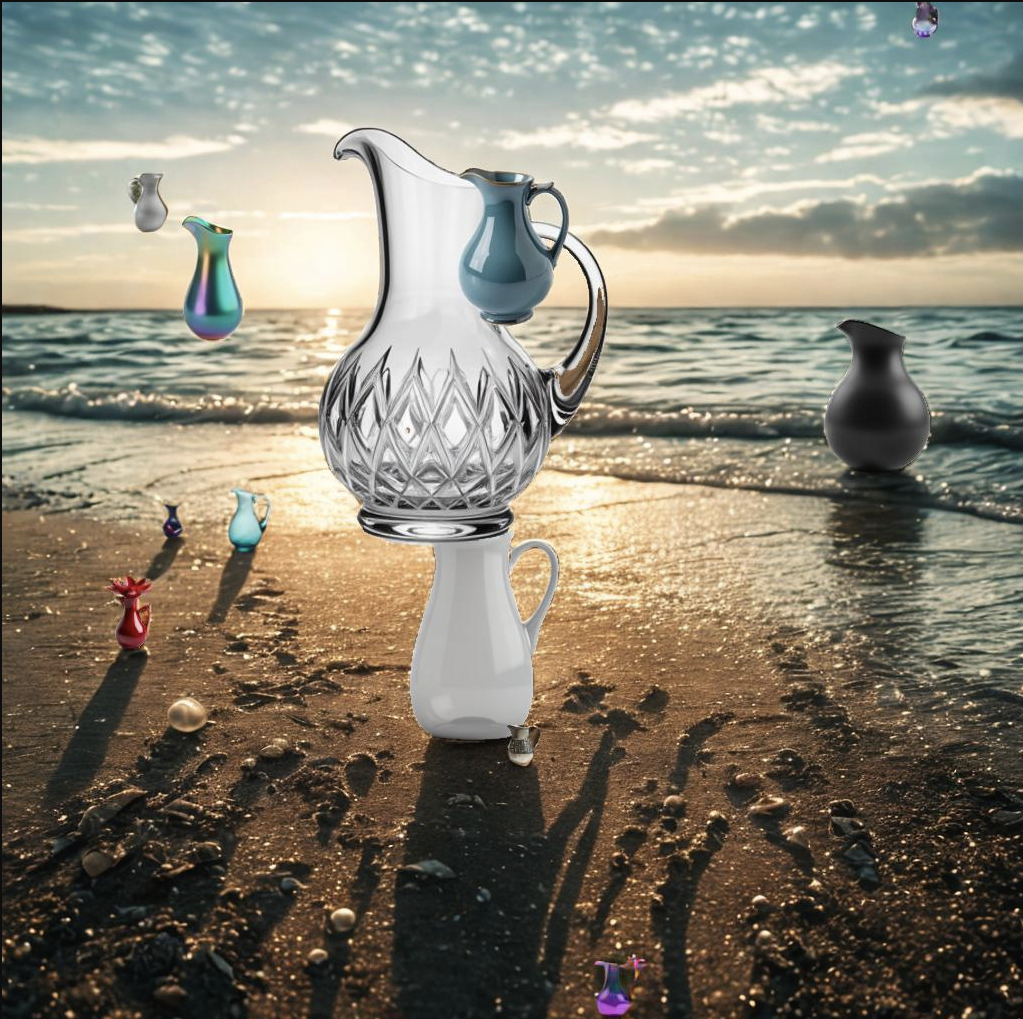}
    \caption{Traininig data sampling from \sfc and \sgc.}
    \label{fig:prompt_referring_expression_our2}
  \end{subfigure}
  \caption{Examples for our target-generated training data.}
  \label{fig:prompt_referring_expression_our}
\end{figure*}

\section{Real Segments Filtering Pipeline}
Our pipeline provides the flexibility of using both synthetic object segments and real object segments collected from a segmentation dataset, We demonstrate that \name pipeline can incorporate real segments from the COCO dataset. Therefore, we build the following real object segments collecting pipeline for future exploration:

Masks extracted directly from the above datasets are frequently occluded, truncated, or loosely bounded, making them unsuitable as composable assets. We therefore introduce a three‑stage pipeline that first filters low‑quality instances and then enriches the survivors with additional metadata.

\textbf{Filtering.}
For every candidate segment we predict three independent quality scores:
\begin{itemize}[left=2pt]
    \item \textbf{Integrity}: Is the object complete and unfragmented?
    \item \textbf{IsObject}: Does the mask depict a discrete ``thing'' (e.g., \emph{car}) rather than amorphous ``stuff'' (e.g., \emph{road})?
    \item \textbf{Mask\,Quality}: How accurately does the mask separate foreground from background?
\end{itemize}
We annotate 4{,}000 samples—with GPT‑4o‐assisted chain‑of‑thought prompts—for ground truth and train three ViT‑B/16 classifiers, one per dimension~\cite{dosovitskiy2021imageworth16x16words}. At inference, we average the predicted scores and retain the top~30\% of segments.
We use this pipeline to filter the SA-1B, COCO, VOC, ADE20K, and get a total of 10M of real object segments for future exploration.


\section{Quality of Synthetic Segments}
Rather than extracting segments from existing photographs, we generate synthetic objects one at a time, ensuring that each segment is complete and free of occlusion. To evaluate quality, we conducted a human‑annotation study in which annotators reviewed 200 randomly sampled segments; 92\% were judged correct.

Because annotators considered the vast majority of synthetic segments high‑quality, we include all of them when composing synthetic images—and have already observed a significant performance boost in downstream models. Additionally, our codebase also includes a lightweight pipeline that uses the CLIP score to measure the semantic similarity between each segment and its caption. We left more exploration on filtering for further work.

\clearpage
\section{Gallery of Object Segments}
\includepdf[pages=1-, scale=0.9, fitpaper=false, pagecommand={}]{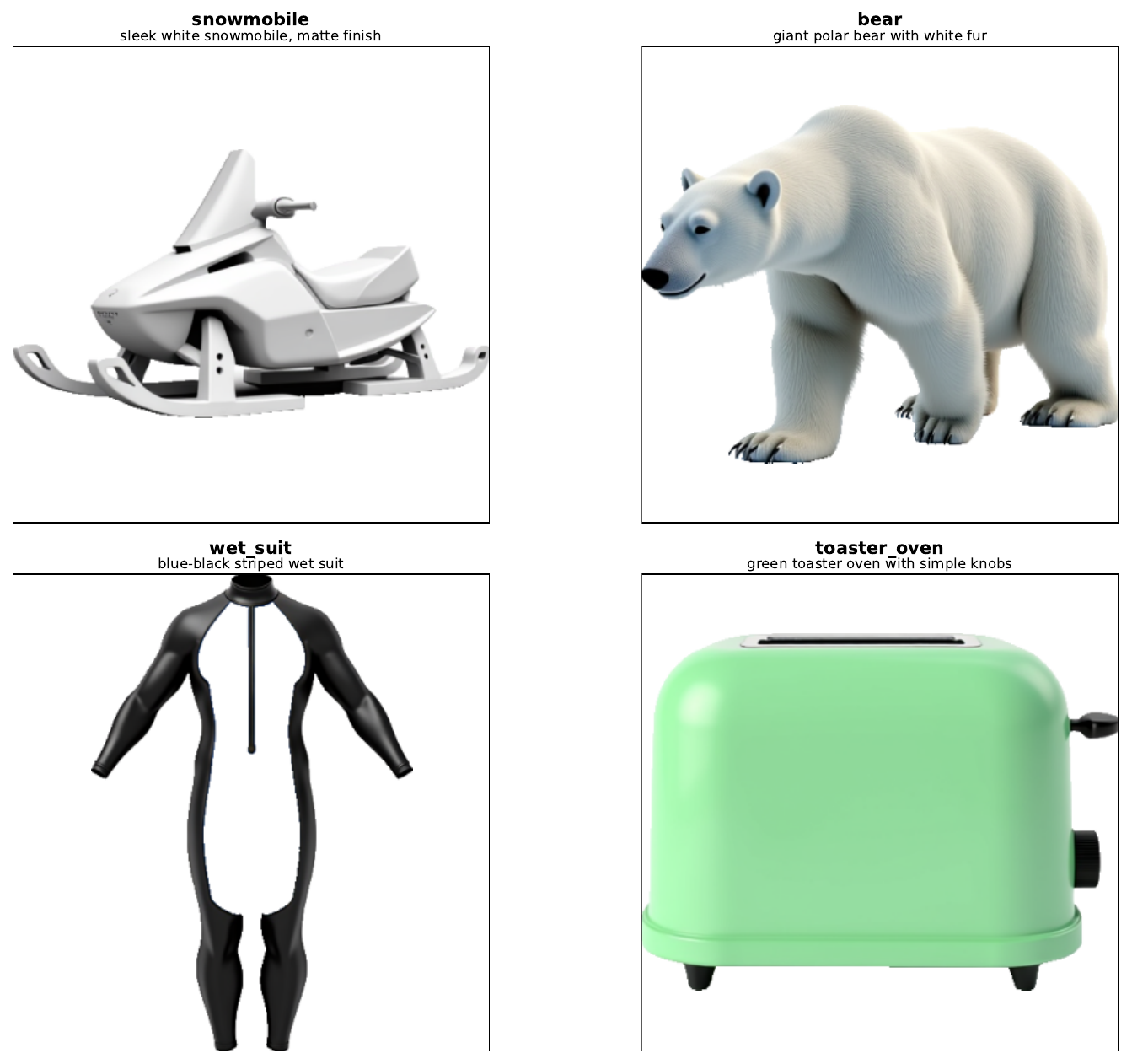}

\clearpage
\section{Gallery of Training Data (Before Camera Configuration Augmentation)}
\noindent\textit{Note: Camera configuration augmentation (random zoom and depth-of-field blur) is applied on-the-fly during training. The images shown here are before this augmentation step.}
\includepdf[pages=1-, fitpaper=false, pagecommand={}]{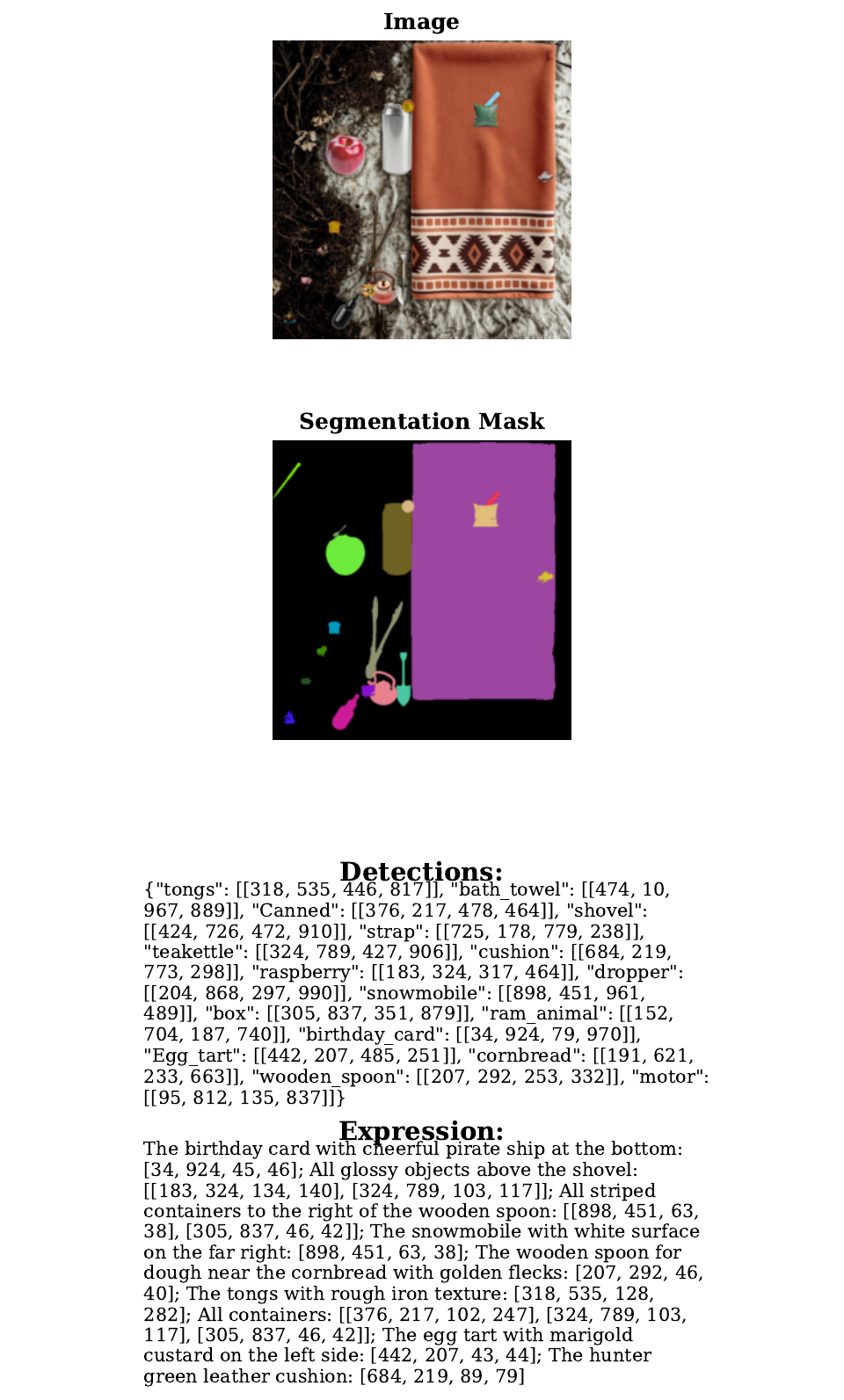}

\clearpage

{
\small
\bibliographystyle{unsrt}
\bibliography{main}
}

\end{document}